\PassOptionsToPackage{normalem}{ulem}
\PassOptionsToPackage{table,dvipsnames}{xcolor}
\ifdefined\pdfminorversion\pdfminorversion=7\fi
\documentclass[]{flare_arxiv_template}

\usepackage[utf8]{inputenc}
\usepackage{times}
\usepackage{latexsym}
\usepackage{inconsolata}
\usepackage{amsmath}
\usepackage{amssymb}
\usepackage{amsfonts}
\usepackage{nicefrac}
\usepackage{makecell}
\usepackage{enumitem}
\usepackage{longtable}
\usepackage{array}
\usepackage{float}
\usepackage{pifont}
\usepackage{xspace}
\usepackage{tabularx}
\usepackage{listings}
\usepackage{tikz}
\usepackage{xfp}
\usepackage{colortbl}
\tcbuselibrary{breakable,skins,listings}

\graphicspath{{image/}{image_v2/}}
\newcolumntype{L}[1]{>{\raggedright\arraybackslash}p{#1}}

\title{FLARE: A Full-Lifecycle Dense Supervision Paradigm for Long-Horizon Coding Agents via Generative Reward Model}
\makeatletter
\renewcommand{\titlelist}{{\titlefont\rmfamily\bfseries\thetitle\par}}
\makeatother

\fancypagestyle{titlestyle}{
  \fancyhf{}
  \fancyhead[L]{\small\sffamily\color{themecolor}\leftmark}
  \fancyhead[R]{\small\sffamily\color{themecolor}FLARE}
  \fancyfoot[C]{\small\sffamily\thepage}
  
  \renewcommand{\headrule}{
  \vskip -2mm
  \hbox to\headwidth{\color{themecolor}\leaders\hrule height 0.5pt\hfill}}
}

\definecolor{deepgreen}{RGB}{0,140,60}
\definecolor{deepred}{RGB}{200,30,30}
\definecolor{grgreen}{RGB}{166,208,168}
\definecolor{blindgray}{RGB}{176,180,186}
\definecolor{radarblue}{RGB}{122,184,210}
\definecolor{grmblue}{RGB}{29,110,170}

\definecolor{codepurple}{RGB}{112,62,145}
\definecolor{codestring}{RGB}{31,94,150}
\lstdefinestyle{experimentxml}{
  language=XML,
  basicstyle=\ttfamily\footnotesize,
  keywordstyle=\color{themecolor}\bfseries,
  identifierstyle=\color{codepurple},
  stringstyle=\color{codestring},
  morekeywords={risk,analysis,issue,severity},
  showstringspaces=false,
  columns=fullflexible,
  keepspaces=true,
  breaklines=true,
  breakatwhitespace=true,
  tabsize=2,
  aboveskip=0pt,
  belowskip=0pt
}
\newtcblisting{experimentcode}{
  enhanced,
  breakable,
  listing only,
  listing engine=listings,
  listing options={style=experimentxml},
  colback=themecolor!3!white,
  colframe=themecolor!60!white,
  colbacktitle=themecolor!12!white,
  coltitle=themedark,
  title={GRM output example},
  fonttitle=\small\sffamily\bfseries,
  boxrule=0.5pt,
  arc=2pt,
  left=7pt,
  right=7pt,
  top=5pt,
  bottom=5pt,
  before skip=7pt,
  after skip=7pt
}

\newcommand{\pos}[1]{\textcolor{deepgreen}{\scriptsize$_{\text{+#1}}$}}

\newcommand{\baddn}[1]{\textcolor{deepred}{\scriptsize$_{\text{-#1}}$}}

\newcommand{\metricbarc}[3]{%
  \edef\ratio{\fpeval{3.2*#1/#2}}%
  \begin{tikzpicture}[baseline=-0.6ex]
    \fill[gray!8] (0,-0.22) rectangle (3.2,0.22);
    \fill[#3!85] (0,-0.22) rectangle (\ratio,0.22);
    \draw[gray!35] (0,-0.22) rectangle (3.2,0.22);
    \node[anchor=west, font=\scriptsize\bfseries, text=black!85]
      at (0.08,0) {\pgfmathprintnumber[1000 sep={,}]{#1}};
  \end{tikzpicture}%
}

\newcommand{\metricbarcB}[3]{%
  \edef\ratio{\fpeval{3.2*#1/#2}}%
  \begin{tikzpicture}[baseline=-0.6ex]
    \fill[gray!8] (0,-0.22) rectangle (3.2,0.22);
    \fill[#3] (0,-0.22) rectangle (\ratio,0.22);
    \draw[gray!35] (0,-0.22) rectangle (3.2,0.22);
    \node[anchor=west, font=\scriptsize\bfseries, text=black!90]
      at (0.08,0) {\pgfmathprintnumber[1000 sep={,}]{#1}};
  \end{tikzpicture}%
}

\makeatletter
\gdef\authorlist{%
  \begin{center}
    \authorfont
    Jingxuan Xu$^{1,*}$\quad
    Gang Wu$^{1,*}$\quad
    Yanan Wu$^{3,*}$\quad
    Yutao Mou$^{2,*}$\\[2pt]
    Songwei Yu$^{1}$\quad
    Tianzhuang He$^{3}$\quad
    Zhengshuo Gong$^{4}$\quad
    Zhao Liu$^{1}$\\[2pt]
    Zihang Xu$^{1}$\quad
    Wenqiang Zhu$^{1}$\quad
    Xinping Lei$^{3}$\quad
    Weihao Li$^{1}$\quad
    Ziqi Miao$^{1}$\\[2pt]
    Yuhui Bai$^{1}$\quad
    Zhongqiu Wang$^{1}$\quad
    Yan Wu$^{1}$\quad
    Ariel Deng$^{1,\dagger}$\\[5pt]
    {\normalfont\small \email{xujingxuan2002@163.com}}
  \end{center}
}
\gdef\affiliationlist{%
  \begin{center}
    \affiliationfont
    $^{1}$Independent Researcher\quad
    $^{2}$Peking University\\
    $^{3}$Nanjing University\quad
    $^{4}$Beijing University of Posts and Telecommunications
  \end{center}
}
\makeatother

\hypersetup{
  pdftitle={FLARE: A Full-Lifecycle Dense Supervision Paradigm for Long-Horizon Coding Agents via Generative Reward Model},
  pdfauthor={Jingxuan Xu, Gang Wu, Yanan Wu, Yutao Mou, Songwei Yu, Tianzhuang He, Zhengshuo Gong, Zhao Liu, Zihang Xu, Wenqiang Zhu, Xinping Lei, Weihao Li, Ziqi Miao, Yuhui Bai, Zhongqiu Wang, Yan Wu, Ariel Deng}
}

\abstract{
While test-time scaling enhances Large Language Model (LLM) agents in long-horizon software engineering (SWE), sparse binary rewards (Pass/Fail) create a severe credit assignment crisis and waste failed exploratory trajectories. Current trajectory optimization and scaling methods are costly and structurally limited, relying on heuristic state reuse without causal diagnosis or delayed scalar scoring without actionable online guidance. We propose FLARE (Full-Lifecycle Alignment and Reward Engine), a novel dense supervision paradigm driven by a lightweight Generative Reward Model (GRM). First, RADAR, an offline causal-aware diagnostic framework, extracts high-fidelity, hindsight-free supervision through causal-chain backtracking to distill a GRM providing real-time, step-level risk feedback. Second, FLARE uses this GRM to continuously optimize the agent across its entire lifecycle. During inference, FLARE acts as an Active Scaffold, autonomously intercepting high-risk generation steps for localized breakpoint re-execution, drastically reducing compute overhead. During post-training, the GRM's structured signals serve as process-supervised reranking scores for Supervised Fine-Tuning (SFT) and step-level dense rewards for Reinforcement Learning (RL), mitigating policy collapse in sparse environments. Extensive evaluations show that FLARE establishes a new Pareto frontier across the agent lifecycle: FLARE (N=1) outperforms Global Rollout (N=5) with a 5x reduction in token consumption. Extending FLARE to training overcomes the sparse reward problem in long-horizon interactive tasks, delivering relative performance gains of 19.13\% in SFT through process-aware data curation and a consistent 9.19\% improvement in RL.
}

\begin{document}
\maketitle
\vspace{-4mm}

{
\renewcommand{\thefootnote}{\fnsymbol{footnote}}
  \footnotetext[1]{Equal contribution.}
  \footnotetext[2]{Corresponding author.}
}

\section{Introduction}

\begin{figure}[t]
    \centering
    \begin{minipage}{0.38\linewidth}
        \centering
        \includegraphics[width=\linewidth]{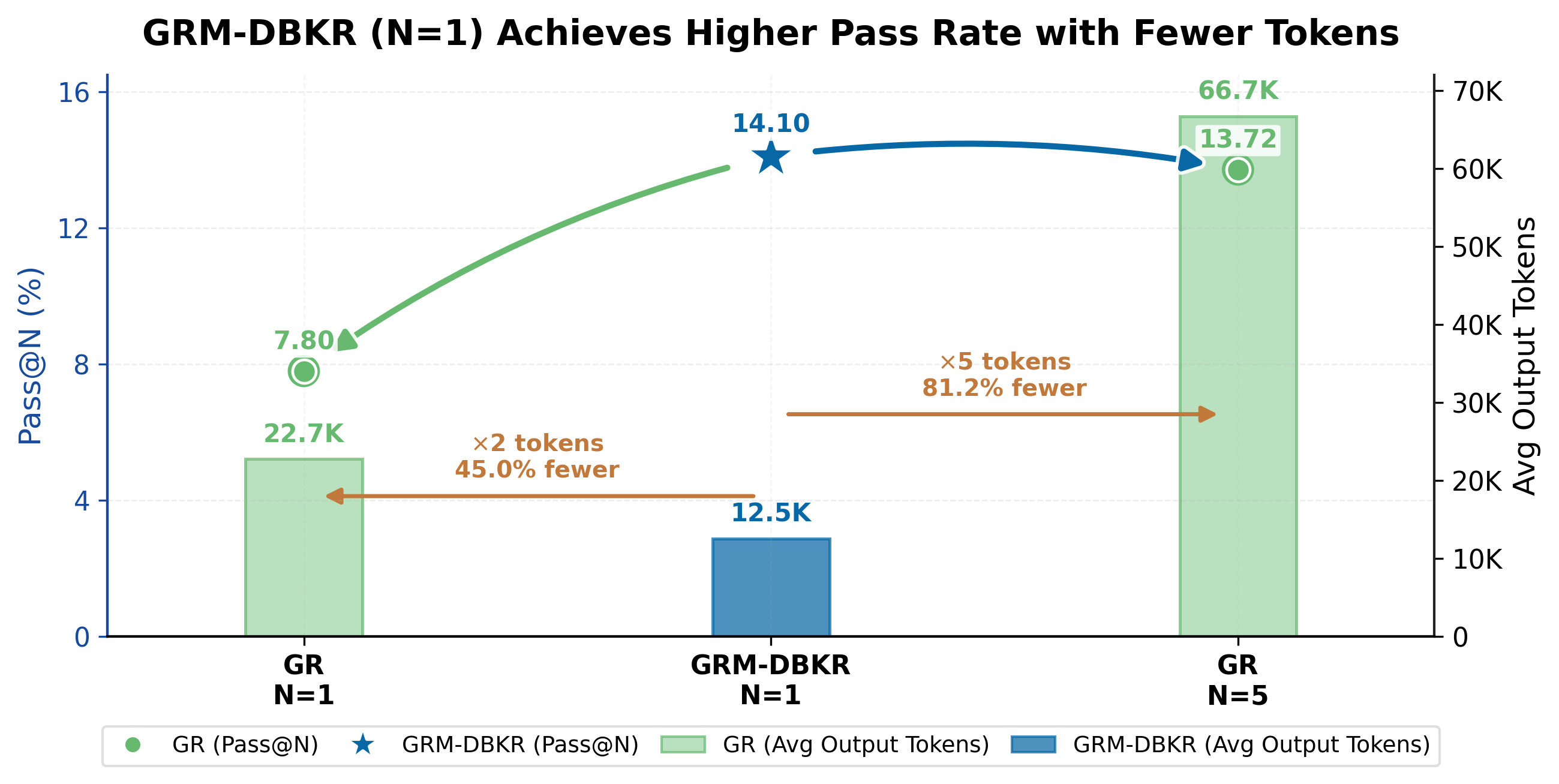}
        \label{fig:intro_grm_dbkr_efficiency}
    \end{minipage}%
    \hspace{0.01\linewidth} %
    \begin{minipage}{0.60\linewidth}
        \centering
        \includegraphics[width=\linewidth]{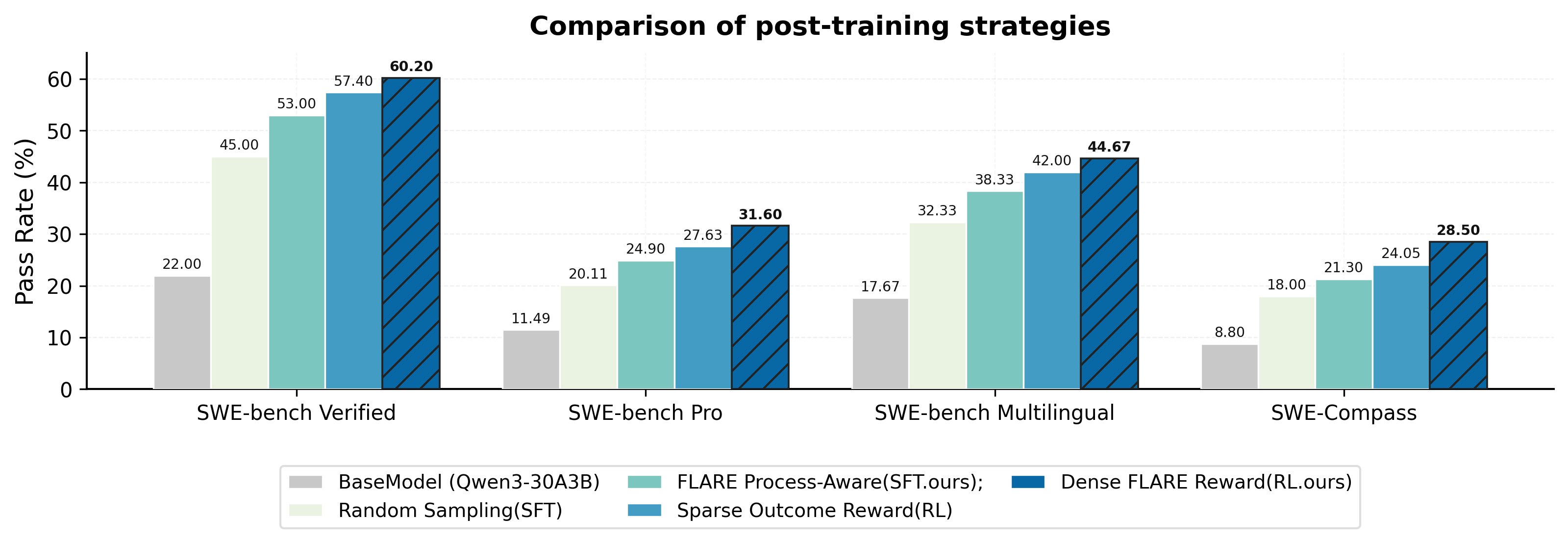}
        \label{fig:rq2_sft_main_comparison}
    \end{minipage}
    \vspace{-0.3cm}
    \caption{
        Comparison of test-time strategies (left) and post-training strategies (right).
    }
    \label{fig:intro}
\end{figure}

Large Language Models (LLMs) have evolved significantly, progressing into sophisticated interactive agents capable of navigating repositories, executing tests, and autonomously resolving complex software engineering (SWE) issues. However, navigating these long-horizon tasks exposes a critical vulnerability: the extreme reliance on sparse, outcome-based binary rewards (Pass/Fail). When an agent fails after executing dozens of steps, a binary outcome offers no insight into the specific point of failure---whether it was a flawed plan at step 5 or a syntax error at step 40. This ``blind trial-and-error'' approach results in a severe credit assignment crisis, forcing the community to discard vast amounts of failed, yet highly informative, exploratory trajectories. 

To enhance capabilities and mitigate these failures, a primary driver of recent progress has been test-time scaling (TTS), which increases inference-time computation to yield higher-quality solutions. Despite its promise, standard TTS is computationally expensive due to the cost of repeated sampling from scratch. Consequently, several recent methods have attempted to optimize this process, yet they face significant structural bottlenecks. Heuristic state reuse methods, such as SWE-Replay~\citep{ding2026swe}, attempt to recycle trajectories by branching at intermediate steps based on structural proxies like reasoning paragraph count. However, without true causal diagnosis, this branching remains a form of heuristic guessing rather than targeted repair. Conversely, offline evolutionary search methods (e.g., Satori-SWE~\citep{zeng2025satori}) and multi-agent debate frameworks (e.g., SWE-Search~\citep{antoniades2024swe}) rely on delayed scalar reward models or heavyweight negotiations. These approaches merely provide trajectory-level scores or incur exponential token overhead, completely failing to intervene dynamically when a trajectory first derails.

To fundamentally break this ``high-cost compute vs. delayed feedback'' deadlock, we introduce \textbf{FLARE (Full-Lifecycle Alignment and Reward Engine)}. FLARE shifts the paradigm from blind global retries and delayed scalar scoring to real-time, lightweight intervention. Our core insight is that true efficiency stems from equipping the agent with self-diagnostic capabilities via a \textbf{Generative Reward Model (GRM)}. To achieve this, we first develop \textbf{RADAR (Root-cause Attribution and Diagnostic Analysis Refiner)}, an offline causal-aware diagnostic framework. Through a dual-track data synthesis pipeline utilizing differential causal-chain backtracking, RADAR extracts precise, hindsight-free supervision signals from complex trajectories. This enables us to train a GRM that is uniquely capable of providing actionable, \textbf{step-level risk feedback}---outputting structured severity levels, error taxonomies, and causal analyses, rather than an uninterpretable scalar score ~\citep{li2026codetracer,mou2026toolsafe}.

Crucially, FLARE is the first paradigm to close the loop across the agent's entire lifecycle. During test-time inference, the GRM operates as an \textit{Active Scaffold}. It asynchronously diagnoses the agent's previously completed step and, when a critical risk is detected, redirects the continuation before the next proposed action is committed. This intervention does not undo an already executed tool operation. Beyond inference, this exact same step-level diagnostic signal is seamlessly internalized during training. FLARE utilizes the GRM's output to curate high-quality demonstrations for Supervised Fine-Tuning (SFT) and constructs step-level dense rewards for Agentic Reinforcement Learning (RL), effectively resolving the sparse-reward bottleneck that plagues long-horizon SWE tasks. 

In summary, we make the following contributions:
\begin{itemize}[leftmargin=0.5cm]
    \item \textbf{The RADAR Framework and GRM Distillation:} We develop an offline diagnostic framework RADAR that utilizes causal-chain backtracking to extract high-fidelity supervision signals, training a lightweight GRM capable of providing structured, step-level diagnostic feedback.
    \item \textbf{A Unified Full-Lifecycle Paradigm:} We introduce FLARE, which integrates the GRM to provide Active Scaffolding for real-time online intervention, while repurposing the same diagnostic signals for SFT data curation and as dense rewards for Agentic RL.
    \item \textbf{Comprehensive Empirical Validation:} We demonstrate that FLARE establishes a new Pareto frontier in test-time scaling: with a strict single-branch budget ($N=1$), it achieves higher success rates than $N=5$ global resampling while achieving a 5x reduction in token consumption. Furthermore, integrating FLARE into the training pipeline yields a 19.13\% relative performance leap in the SFT stage, and a 9.19\% average relative improvement  during RL across four major SWE benchmarks, as illustrated in Figure \ref{fig:intro}.
\end{itemize}

\section{Related Work}
\paragraph{Test-Time Scaling and Trajectory Optimization in Long-Horizon Interactive Tasks.} 
Repository-level evaluation covers test-grounded implementation~\citep{yu2026swenexus} and trajectory-level engineering constraints~\citep{zhu2026anchor}. Coding agents such as SWE-agent, AutoCodeRover, and Agentless~\citep{yang2024swe,zhang2024autocoderover,xia2024agentless} have made test-time scaling (TTS) a central mechanism for improving long-horizon software engineering performance. Representative TTS optimizations include \textit{heuristic state reuse}, exemplified by SWE-Replay~\citep{ding2026swe}, which resumes exploration from intermediate states using lightweight heuristics, and \textit{offline evolutionary search}, represented by Satori-SWE~\citep{zeng2025satori}, which refines complete trajectories using delayed feedback. These approaches improve sample efficiency through state reuse or trajectory refinement. FLARE instead couples branch selection with step-level diagnosis: an online Generative Reward Model (GRM) identifies critical deviations and provides targeted feedback for Active Scaffolding.

\paragraph{Trace Diagnosis and Failure Attribution.}
A complementary line of research focuses on diagnosing why agents fail in long-horizon environments. Frameworks like CodeTracer~\citep{li2026codetracer} make agent states explicitly traceable, while TRAIL, AgenTracer, and TRACER~\citep{deshpande2025trail,zhang2025agentracer,tayebati2026tracer} study error localization, failure attribution, and trajectory-level risk. Other work identifies the responsible agent or decisive step in multi-agent failures~\citep{zhangagent}, while code-critique benchmarks assess diagnostic feedback through fine-grained checklists~\citep{zhang2025codecriticbench}. CodeTracer further demonstrates that localized diagnostic evidence can support replay-based recovery. FLARE advances this diagnostic perspective through RADAR, which converts offline causal-chain analysis into prefix-grounded supervision for a lightweight online GRM. Its structured severity, error taxonomy, and repair rationale are directly reusable for both breakpoint intervention and policy training.
\paragraph{Process Supervision and Critical-Step Learning.}
To combat reward sparsity, Process Reward Models (PRMs) have been widely adopted to evaluate intermediate reasoning steps, traditionally in mathematical or multimodal reasoning~\citep{lightman2023let,zhangprogress}. Complementary credit-assignment schemes use dependency-aware test credit in code tasks~\citep{li2026weave} or role-specific downstream utility in multimodal reasoning~\citep{miao-etal-2026-seeing}. In the agentic SWE domain, methods like STeCa and ATLaS~\citep{wang2025steca,chen2025atlas} demonstrate that extracting critical steps can outperform full-trajectory behavior cloning. Scalar process scores summarize the quality of intermediate decisions, whereas textual diagnostics additionally expose the failure mechanism and corrective direction. FLARE adopts a conditional generative formulation: the GRM produces structured step-level feedback that supports actionable intervention and is subsequently converted into dense rewards for Train-Time Alignment.

Related approaches also reuse interpretable process feedback. SWE-PRM delivers taxonomy-guided feedback during execution~\citep{gandhi2025astray}; rubric-supervised critics support trajectory selection and data curation~\citep{wang2026rubric}; and SWE-TRACE combines trajectory construction, process-guided RL, and inference-time guidance~\citep{han2026swetrace}. Feedback is also internalized through reminder-conditioned on-policy distillation~\citep{luo2026oprd} or visual-critic-guided RL with critic-free deployment~\citep{li2025relook}, while multi-stage coding curricula combine supervised and reinforcement fine-tuning~\citep{zhan2025katcoder}. FLARE centers this lifecycle on a shared diagnostic representation learned from RADAR's failed-trajectory backtracking and active error injection. The same representation specifies when to intervene, what to repair, and how to rank and reward trajectories, connecting causal-aware supervision construction to all three downstream uses.

\section{Method}

\begin{figure*}[t]
    \centering
    \resizebox{1.0\linewidth}{!}{
    \includegraphics{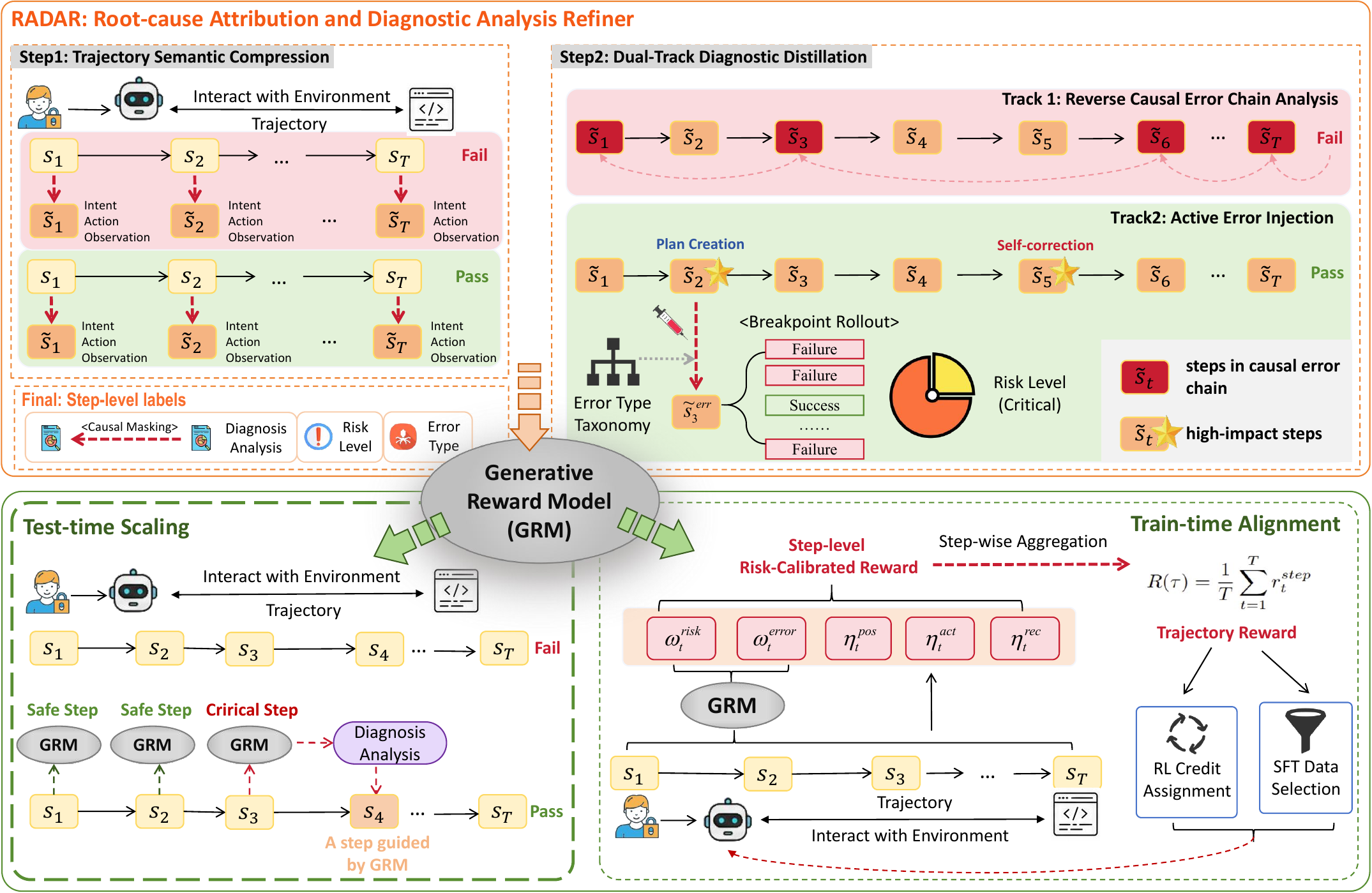}}
    \caption{\textbf{FLARE overview.} RADAR extracts step-level supervision signals from agent trajectories for training a lightweight Generative Reward Model (GRM), which provides real-time step-wise feedback during execution, enabling both test-time scaling and training-time alignment.}
    \label{fig:method}
\end{figure*}

To address reward sparsity~\citep{wei2025swe} and delayed post-hoc feedback~\citep{li2026codetracer} in long-horizon SWE tasks, we propose 
\textbf{FLARE} (\textbf{F}ull-\textbf{L}ifecycle \textbf{A}lignment and \textbf{R}eward \textbf{E}ngine), a novel dense supervision paradigm driven by a lightweight \emph{Generative Reward Model} (GRM) that provides real-time step-level supervision during agent execution (as shown in Figure \ref{fig:method}).
First, we propose \textbf{R}oot-cause \textbf{A}ttribution and \textbf{D}iagnostic \textbf{A}nalysis \textbf{R}efiner (\textbf{RADAR}), which extracts fine-grained, step-level supervision signals from agent trajectories for GRM optimization (Section \ref{sec:radar}). The GRM then provides real-time, step-wise supervision during SWE agent execution, enabling both test-time  and training-time alignment (Section \ref{sec:grm}).

\subsection{RADAR for GRM Optimization: From Trajectory to Step-wise Supervision}
\label{sec:radar}

The key challenge in training a Generative Reward Model (GRM) lies in obtaining high-quality, causally consistent step-level supervision from agent trajectories, where only a small subset of steps truly leads to the final failure. To address this, we propose a unified framework in which \textbf{RADAR} serves as a data synthesis engine. Given a trajectory $\tau = (s_1, \dots, s_T)$ for a task $q$, RADAR identifies causally relevant steps and assigns structured step-wise annotations, thereby producing fine-grained supervision signals. These step-level annotations are then used to train GRM as a conditional generative model that produces structured reward-aligned diagnostic outputs at each step.

\subsubsection{RADAR as a Step-level Supervision Engine}

RADAR extracts step-level supervision signals from agent trajectories to support GRM optimization. It first performs trajectory semantic compression by summarizing each step and retaining only the key information. It then applies a dual-track diagnostic distillation process: for failed trajectories, RADAR conducts reverse causal error-chain analysis to identify the risk level, error type, and diagnostic analysis of each step; for successful trajectories, it introduces active error injection to augment long-tail error-type samples and improve supervision diversity.

\paragraph{Step1: Trajectory Semantic Compression.}
Each step $s_t$ is first compressed into a structured semantic representation
$\tilde{s}_t = (I_t, A_t, O_t) = \phi(s_t)$, 
where $I_t$, $A_t$, and $O_t$ denote the intent, action, and observation, respectively, all represented as natural language summaries. This abstraction provides a unified basis for subsequent dual-track diagnostic distillation.

\paragraph{Step2: Dual-track Diagnostic Distillation}

\begin{itemize}[leftmargin=0.5cm]
    \item \textbf{Track 1: Reverse Causal Error Chain Analysis.}
    For failed trajectories, RADAR identifies a root-prioritized set of causally relevant steps via backward tracing:
    \begin{equation}
    \mathcal{C} = \mathcal{B}(\tau; \mathcal{G}) = (\tilde{s}_{t_1}, \dots, \tilde{s}_{t_k})
    \end{equation} 
    where $\mathcal{G}$ contains global signals (e.g., patch differences and failed test cases), and each $\tilde{s}_{t_i}$ is retained through an evidence-supported dependency on the observed failure. RADAR first anchors a behavioral or contract discrepancy, then traces backward through steps that introduce, propagate, or reinforce it. The earliest supported source is prioritized for replay; unrelated exploration is excluded, while independent defects may retain separate roots. This causal-aware procedure targets explanatory error chains rather than a formally identified minimal causal set.
    Each step in the causal chain is assigned a structured label: 
    \begin{equation}
    y_{t_i} = (z_{t_i}, c_{t_i}, d_{t_i}) = \mathcal{A}(\tilde{s}_{t_i}; \mathcal{C}, \mathcal{G})
    \end{equation}
    where $\mathcal{A}$ denotes the annotation function conditioned on the full causal chain $\mathcal{C}$ and global signals $\mathcal{G}$. Here, $z_{t_i}$ denotes the risk level, $c_{t_i}$ denotes the error type under a predefined taxonomy (see Appendix \ref{app:taxonomy} for details), and $d_{t_i}$ is an offline diagnostic analysis derived from the global signals.

    \item \textbf{Track 2: Active Error Injection.}
    To improve error diversity and balance, RADAR perturbs successful trajectories by injecting errors into predefined high-impact steps, including \textit{Plan Creation}, \textit{Critical Observation}, \textit{Critical Action}, and \textit{Self-Correction}. Key steps are selected as:
    \begin{equation}
    \mathcal{K} = \{ \tilde{s}_t \mid s_t \in \tau^{+},\; \kappa(\tilde{s}_t) \in \mathcal{H} \},
    \end{equation}
    where $\kappa(\cdot)$ denotes a step-type classifier and $\mathcal{H}$ is the predefined high-impact category set. Unlike semantic compression $\phi$, $\kappa$ assigns a functional role to each compressed step. For each selected step, RADAR samples an error label $\ell \sim p(\ell)$ and applies a label-specific perturbation policy:
    $\tilde{s}_t^{\text{err}} = \mathcal{I}(\tilde{s}_t \mid \ell, \pi_\ell)$. The modified trajectory is then re-executed via breakpoint rollout, producing stochastic rollouts $\{\mathcal{T}_t^{(n)}\}_{n=1}^N$ with empirical recovery rate:
    $
    \hat{p}_t = \frac{1}{N} \sum_{n=1}^{N} \mathbb{I}(\mathcal{T}_t^{(n)} \text{ solves task}).
    $
    Injected errors are finally categorized using a recovery threshold $\theta \in [0,1]$:
    \begin{equation}
    z_t =
    \begin{cases}
    \textit{critical}, & \hat{p}_t \leq \theta, \\
    \textit{trivial}, & \hat{p}_t > \theta.
    \end{cases}
    \end{equation}
\end{itemize}

The two tracks provide complementary supervision: natural failures supply observed error chains, while perturbations of successful prefixes expose underrepresented failure modes. Track 2 severity is grounded in observed recoverability under the continuation policy and rollout budget, rather than assigned solely from the injected error type. The augmentation targets coverage, not the natural frequency of errors.

\paragraph{Final: Step-level Labels}

Through the above process, each trajectory $\tau^{(i)}$ is converted into a sequence of annotated tuples:
\begin{equation}
\{(\tilde{s}_t^{(i)}, y_t^{(i)})\}^T_{t=1}
\end{equation}
forming a large-scale dataset with fine-grained step-wise supervision signals. Causal-prefix distillation separates the evidence used to construct labels from the evidence available to an online explanation:
\[
\tilde{d}_t = \mathcal{M}(d_t; q,s_t,\mathcal{H}_t), \qquad
\mathcal{H}_t = (s_1,\dots,s_{t-1}).
\]
Here, $\mathcal{M}$ rewrites the offline analysis using only the task, current step, and preceding history. Reference patches and final outcomes may inform offline attribution, but they are not available to the runtime GRM and must not appear as prefix-invisible evidence in $\tilde{d}_t$. A semantic compliance gate checks taxonomy consistency, unsupported anchors, and patch-, future-, or test-oracle leakage, routing violations for regeneration. The resulting target is $\tilde{y}_t=(z_t,c_t,\tilde{d}_t)$; the full offline analysis is not used directly. This defines the operational information boundary of hindsight-free supervision without asserting a formal zero-leakage guarantee. The annotation, distillation, and compliance prompts are provided in Appendix~\ref{app:radar_prompts}.

\subsubsection{GRM Optimization with Step-wise Supervision}

We train GRM as a conditional generative model:
\begin{equation}
p_{\psi}(\tilde{y}_t \mid q,s_t,\mathcal{H}_t),
\end{equation}
where $\psi$ denotes the GRM parameters and $\mathcal{H}_t$ is the preceding interaction history.
Given the RADAR-constructed dataset, GRM is optimized to generate structured outputs including risk level, error type, and causal-consistent diagnostic explanations.

In practice, we perform supervised fine-tuning on Qwen3.5-4B, enabling a lightweight GRM to produce step-wise risk assessment and interpretable error analysis conditioned on trajectory context. The actor and critic solve different problems: the actor searches for and implements a repair, whereas the GRM diagnoses a completed step against the task, available evidence, and a fixed error taxonomy. Structured supervision concentrates the smaller model on this diagnostic role without requiring it to solve the entire issue independently. The cross-model and cross-scaffold results in Section~\ref{sec:rq1} support this complementary use.

\subsection{GRM for Test-Time Intervention and Training-Time Alignment}
\label{sec:grm}

Our GRM provides real-time step-level supervision during SWE agent execution, enabling both test-time intervention and train-time alignment. Its shared diagnostic representation serves three complementary functions: severity determines when to intervene, the diagnostic rationale specifies what to repair, and severity together with taxonomy information defines the process score used for SFT selection and dense reward construction. The following mechanisms reuse one diagnostic model rather than requiring a separate evaluator for each stage.

\subsubsection{Test-Time Scaling.}
At each step, the agent proposes a tentative action $\hat{s}_t$ while GRM asynchronously diagnoses the previously completed step $s_{t-1}$, producing a diagnostic triplet $y_{t-1} = (z_{t-1}, c_{t-1}, d_{t-1}^{\mathrm{diag}})$. The intervention therefore acts at the boundary before the next action is committed; it does not assume that an already executed tool operation can be undone. If the diagnosis indicates a non-critical state, the agent proceeds with $\hat{s}_t$; otherwise, it regenerates the next step conditioned on the diagnostic feedback:
\begin{equation}
s_t =
\begin{cases}
\hat{s}_t, & z_{t-1} \in \{\textit{safe}, \textit{trivial}\}, \\
\pi(d_{t-1}^{\mathrm{diag}};\mathcal{H}_t), & z_{t-1} = \textit{critical},
\end{cases}
\end{equation}
where $\pi(\cdot)$ denotes regeneration conditioned on diagnostic feedback and interaction history.

\subsubsection{Train-Time Alignment.}

To optimize the agent with fine-grained supervision, we formulate a step-level risk-calibrated reward function that captures diagnostic severity and recoverability, based on empirical statistics derived from breakpoint rollouts on offline GRM-annotated trajectories. This connects reward construction to executable continuation outcomes rather than assigning fixed penalties solely from taxonomy labels.

First, let $\bar{P}_{\text{succ}}$ and $\bar{P}_{\text{fail}}$ denote the global baseline success and failure rates, respectively, obtained by randomly sampling steps from offline trajectories for breakpoint rollout. For the $t$-th step in a trajectory, let $z_t \in \{\text{safe}, \text{trivial}, \text{critical}\}$ denote its risk severity level, and $c_t$ denote the specific error type triggered. We define the \textbf{Risk Severity Factor} $\omega^{\text{risk}}_t$ and the \textbf{Error Importance Factor} $\omega^{\text{err}}_t$:
\begin{equation}
\omega^{\text{risk}}_t = 
\begin{cases} 
\max \left( 0, \frac{P_{\text{succ}}(z_t) - \bar{P}_{\text{succ}}}{\bar{P}_{\text{succ}}} \right), & \text{if } z_t = \text{safe}, \\
\max \left( 0, \frac{P_{\text{fail}}(z_t) - \bar{P}_{\text{fail}}}{\bar{P}_{\text{fail}}} \right), & \text{if } z_t \in \{\text{trivial}, \text{critical}\},
\end{cases}
\end{equation}
\begin{equation}
\omega^{\text{err}}_t = \max \left( 0, \frac{\bar{P}_{\text{succ}} - P_{\text{succ}}(c_t)}{\bar{P}_{\text{succ}}} \right),
\end{equation}
where $P_{\text{succ}}(\cdot)$ and $P_{\text{fail}}(\cdot)$ denote the empirically estimated success and failure rates, respectively, precomputed from offline breakpoint rollouts conditioned on a specific risk severity level or error type. These factors effectively translate offline statistical patterns into reward or penalty weights for critical steps.

Combining these statistical priors with the dynamic characteristics of the trajectory, we formalize the step-level risk-calibrated reward $r^{step}_t$ for step $t$ as:
\begin{equation}
r^{step}_t =
\begin{cases}
\omega^{\text{risk}}_t \cdot \eta^{\text{pos}}_t \cdot \eta^{\text{act}}_t, & \text{if } z_t = \text{safe}, \\
- \omega^{\text{risk}}_t \cdot \omega^{\text{err}}_t \cdot \eta^{\text{pos}}_t \cdot \eta^{\text{act}}_t \cdot \eta^{\text{rec}}_t, & \text{if } z_t \in \{\text{trivial}, \text{critical}\}.
\end{cases}
\end{equation}
where three dynamic adjustment coefficients tailored to the current trajectory state are defined as follows:
\begin{itemize}[leftmargin=0.5cm]
    \item \textbf{Position-aware weight} $\eta^{\text{pos}}_t = \frac{1 + \log t}{1 + \log T}$: Emphasizes proximity to the submitted solution while retaining positive weight for early decisions. Logarithmic growth limits concentration on the final steps, and normalization sets the terminal weight to one ($T$ is the trajectory length).
    \item \textbf{Action rarity weight} $\eta^{\text{act}}_t = 1 - \frac{N(a_t)}{T}$: Downweights reward contributions from high-frequency actions, where $N(a_t)$ is the occurrence count of action $a_t$. Because this factor multiplies both positive and negative rewards, it attenuates their magnitudes rather than imposing an additional negative penalty.
    \item \textbf{Recurrence penalty} $\eta^{\text{rec}}_t=1+\frac{n_t^{(c)}\left(n_t^{(c)}-1\right)}{2}$, where $n_t^{(c)}$ counts occurrences of error type $c_t$ up to step $t$. The pair-count term increases the penalty as an error recurs, targeting repeated failures rather than repetition of an action alone.
\end{itemize}

This step-level formulation yields a dense reward that jointly captures diagnostic severity, temporal position, action frequency, and error recurrence. The modifiers encode explicit preferences for trajectory quality rather than estimates of individual causal effects. Their utility is supported by the component ablations in Section~\ref{sec:rq2}, where severity and error importance provide the largest gains and the remaining terms refine the score.
We then define the trajectory-level reward as the normalized aggregation of step-level signals:
\begin{equation}
R(\tau) = \frac{1}{T} \sum_{t=1}^{T} r^{step}_t,
\end{equation}
where $T$ is the trajectory length and $r^{step}_t$ is the GRM-derived step reward. This trajectory-level score provides a unified criterion for evaluating the overall quality of generated rollouts, and is further used to guide both supervised and reinforcement learning stages.

\paragraph{SFT Data Selection.}
To improve SFT quality, we move away from random trajectory sampling and instead select high-quality demonstrations based on the trajectory-level score $R(\tau)$. Specifically, we retain the top-ranked subset of trajectories with the highest $R(\tau)$, enabling the model to better internalize stable reasoning patterns and self-correction behaviors. 

\paragraph{Dense Reward Integration.}

Prior work optimizes a sparse trajectory-level signal, assigning each trajectory $\tau^{(i)}$ a binary pass/fail reward $R^{out}(\tau^{(i)}) \in \{0,1\}$. We extend this by incorporating a reranking reward that aggregates step-level supervision signals, yielding a unified trajectory reward $r^{(i)} = R(\tau^{(i)})+R^{out}(\tau^{(i)})$.

Within each group, we compute relative advantages via z-score normalization:
\[
\hat{A}_i = \frac{r^{(i)} - \mu_G}{\sigma_G + \epsilon}, \quad
\mu_G = \frac{1}{G}\sum_j r^{(j)}, \quad
\sigma_G = \sqrt{\frac{1}{G}\sum_j (r^{(j)} - \mu_G)^2}.
\]
All tokens in $\tau^{(i)}$ share the same $\hat{A}_i$. We apply standard GRPO-based optimization~\citep{shao2024deepseekmath}, using a trajectory-level advantage to integrate the dense step rewards. Here, \textit{dense} refers to supervision constructed throughout the trajectory, not to a separate advantage for every step or token. Trajectories with the same terminal Pass/Fail outcome can receive different rewards through $R(\tau)$, allowing policy optimization to distinguish their process quality even after aggregation.

\newcommand{\auditmark}{%
  \leavevmode\pdfliteral direct{/Span << /ActualText <FEFFFFE5> >> BDC}%
  \makebox[1em][c]{\textyen}%
  \pdfliteral direct{EMC}%
}
\expandafter\def\csname u8:\detokenize{}\endcsname{\auditmark}

\section{Experiments}
\label{sec:experiments}

We evaluate FLARE along three research questions, each corresponding to one use of the GRM across the agent lifecycle:
\begin{itemize}[leftmargin=0.5cm]
    \item \textbf{RQ1: Test-Time Intervention and Inference Efficiency.} Can the GRM halt and repair failing trajectories online more effectively than global resampling or post-hoc guided replay?
    \item \textbf{RQ2: Process-Aware Trajectory Scoring and SFT Data Curation.} Does the GRM-derived trajectory score identify higher-quality demonstrations for SFT than random or rubric-critic selection?
    \item \textbf{RQ3: Dense Reward for Agentic Reinforcement Learning.} Can GRM-derived dense rewards improve long-horizon SWE policy optimization over outcome-only sparse rewards?
\end{itemize}

\subsection{Experimental Setup}
\label{sec:exp_setup}

\paragraph{Benchmarks and agents.}
We evaluate FLARE on four repository-level SWE benchmarks: SWE-bench Verified, SWE-bench Pro~\citep{deng2025swepro}, SWE-bench Multilingual~\citep{swebenchmultilingual}, and SWE-Compass~\citep{xu2025swe}. These benchmarks cover issue-resolution tasks with different difficulty levels, repository structures, and programming languages. RQ1 covers GLM-5, Kimi K2.5, Qwen3.5-Plus, and DeepSeek V3.2~\citep{glm52025,kimi2025k2,qwen352026,deepseek2025v3}, deployed with Claude Code and Cline-style scaffolds~\citep{claudecode2024,cline2024}; aggregate and disaggregated results are distinguished below. RQ2 fine-tunes Qwen3-30B-A3B~\citep{qwen32025}. Unless otherwise specified, comparisons within each experiment use the same repository-level test harness, task prompts, branch budgets, and base-agent configurations.

\paragraph{Test-time intervention baselines.}
For RQ1, we compare four repair strategies that isolate the value of full reruns, checkpoint reuse, offline diagnosis, and online GRM intervention:
\begin{itemize}[leftmargin=0.5cm]
    \item \textbf{GR} (\textit{Global Rollout}) is the standard Pass@$N$ baseline: for each task, the agent samples $N$ complete trajectories from the initial task state, and the task is counted as solved if any trajectory passes. GR does not use checkpoints, branching, or reward-model feedback.
    \item \textbf{Blind-BKR} (\textit{Blind Breakpoint Rollout}) isolates the effect of state reuse. After an initial failure, the environment is restored to the first high-risk step, and the agent continues from that checkpoint without diagnostic guidance.
    \item \textbf{RADAR-DBKR} (\textit{RADAR diagnostic-guided breakpoint rollout}) is a post-hoc diagnostic replay baseline: it waits for a complete failed trajectory, truncates the trajectory at the root-cause step identified by RADAR, appends RADAR diagnostic feedback as user content, and performs a localized breakpoint rollout.
    \item \textbf{GRM-DBKR} (\textit{Online GRM diagnostic-guided breakpoint rollout}) is the online FLARE intervention method. The GRM diagnoses completed steps and, when critical risk is detected, provides localized diagnostic feedback to regenerate the next action before it is committed.
\end{itemize}

\paragraph{Budget protocol.}
For a repair strategy, $N$ denotes the number of repair branches allowed for each task. For Global Rollout, these branches are full trajectories sampled from the initial state. For checkpoint-based methods, they start from the selected or detected checkpoint. A task is counted as repaired if at least one branch produces a patch that passes the repository tests. Costs include all budgeted branches, not only the successful branch. Under the reported conditional protocol, GR, Blind-BKR, and RADAR-DBKR include the initial baseline rollout, while GRM-DBKR counts its online-guided rollouts. Thus, matching $N$ controls branch count, not the number of full trajectories or model calls.

\paragraph{Trajectory data and training comparisons.}
The SFT and RL data pools are built from \textit{multi-swe-rl}, \textit{rebench}, and supplementary GitHub instances. For trajectory scoring and SFT selection, we generate 18,000 complete inference trajectories via instance distillation, including 6,379 passing trajectories and 11,621 failing trajectories. For RQ2, all SFT data-selection methods choose 3K trajectories from this pool and fine-tune Qwen3-30B-A3B with the same optimization recipe; only the selection signal changes. We compare \textbf{Random Sampling}, \textbf{Rubric-Supervised Critic Selection}, and \textbf{FLARE Process-Aware Selection}. The rubric-critic baseline follows \citet{wang2026rubric}, which learns a critic from sparse real-world outcomes using rubric-based supervision. It is distinct from the heuristic \textbf{Critical Model} baseline in the ROC-AUC comparison. For RQ3, all RL variants start from the same initialization and use the same training data source, rollout budget, optimizer, learning-rate schedule, and evaluation harness; only the reward signal differs.

\paragraph{Metrics.}
For RQ1, we report fail-to-pass (F2P) rate on initially failed tasks and pass-to-pass (P2P) rate on initially successful tasks. F2P uses Pass@$N$, where a task succeeds if any of the $N$ branches passes. P2P uses Avg@$N$, the mean pass rate across branches, to assess preservation of successful behavior. These task/trajectory-level metrics are distinct from the F2P/P2P test-case labels used by the SWE-bench harness. We also report average base-agent output tokens and agent turns under the budget protocol; token counts exclude agent inputs and GRM inputs/outputs. For RQ2 and RQ3, the downstream metric is benchmark pass rate, also referred to as resolve rate. The main tables report point estimates.

For training-free trajectory scoring in RQ2, we report ROC-AUC. Given a trajectory score $R(\tau)$ and binary outcome label $y \in \{0,1\}$, where $y=1$ denotes a passing trajectory, let $\mathcal{P}=\{i:y_i=1\}$ and $\mathcal{N}=\{j:y_j=0\}$. ROC-AUC is computed as:
\begin{equation}
\mathrm{AUC} =
\frac{1}{|\mathcal{P}||\mathcal{N}|}
\sum_{i\in\mathcal{P}}\sum_{j\in\mathcal{N}}
\left[
\mathbb{I}\!\left(R(\tau_i)>R(\tau_j)\right)
+
\frac{1}{2}\mathbb{I}\!\left(R(\tau_i)=R(\tau_j)\right)
\right].
\label{eq:roc_auc}
\end{equation}
This metric measures the probability that a randomly sampled passing trajectory is ranked above a randomly sampled failing one, with ties receiving half credit. It is appropriate because the GRM score is used as a ranking signal for top-$K$ data selection rather than as a calibrated success probability.

\paragraph{GRM output format.}
The GRM is trained and used with a structured XML output contract so that its signal can be reused for online repair, trajectory scoring, and dense reward construction. For risky steps, the model outputs a causal analysis, one or more taxonomy-backed issue tags, and a severity label:
\begin{experimentcode}
<risk>
  <analysis>
    [Detailed causal analysis and actionable repair advice]
  </analysis>
  <issue rubric="E1.1" label="Misunderstood Requirements"/>
  <issue rubric="E1.4" label="Wrong Tool Selection"/>
  <severity>critical</severity>
</risk>
\end{experimentcode}
This structured output prevents the GRM from acting as a monolithic scalar reranker and instead exposes severity, taxonomy, and repair guidance for downstream use.

\subsection{RQ1: Test-Time Repair and Inference Efficiency}
\label{sec:rq1}

\begin{figure}[t]
    \centering
    \begin{minipage}{0.48\linewidth}
        \centering
        \includegraphics[width=\linewidth]{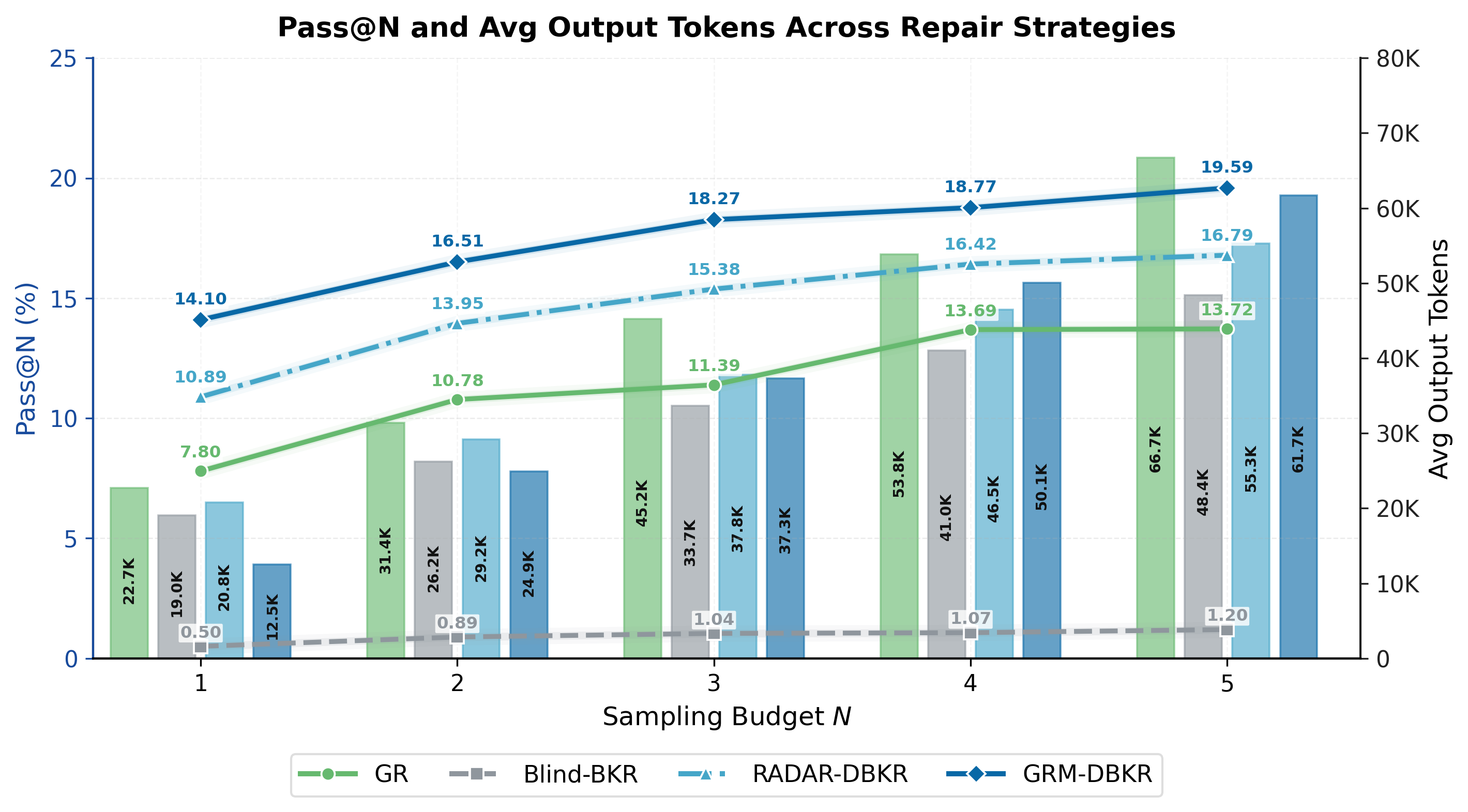}
        \caption{
            F2P repair curves under increasing branch budgets, aggregated across RQ1 model--scaffold settings.
        }
        \label{fig:f2p_curve}
    \end{minipage}
    \hfill
    \begin{minipage}{0.48\linewidth}
        \centering
        \includegraphics[width=\linewidth]{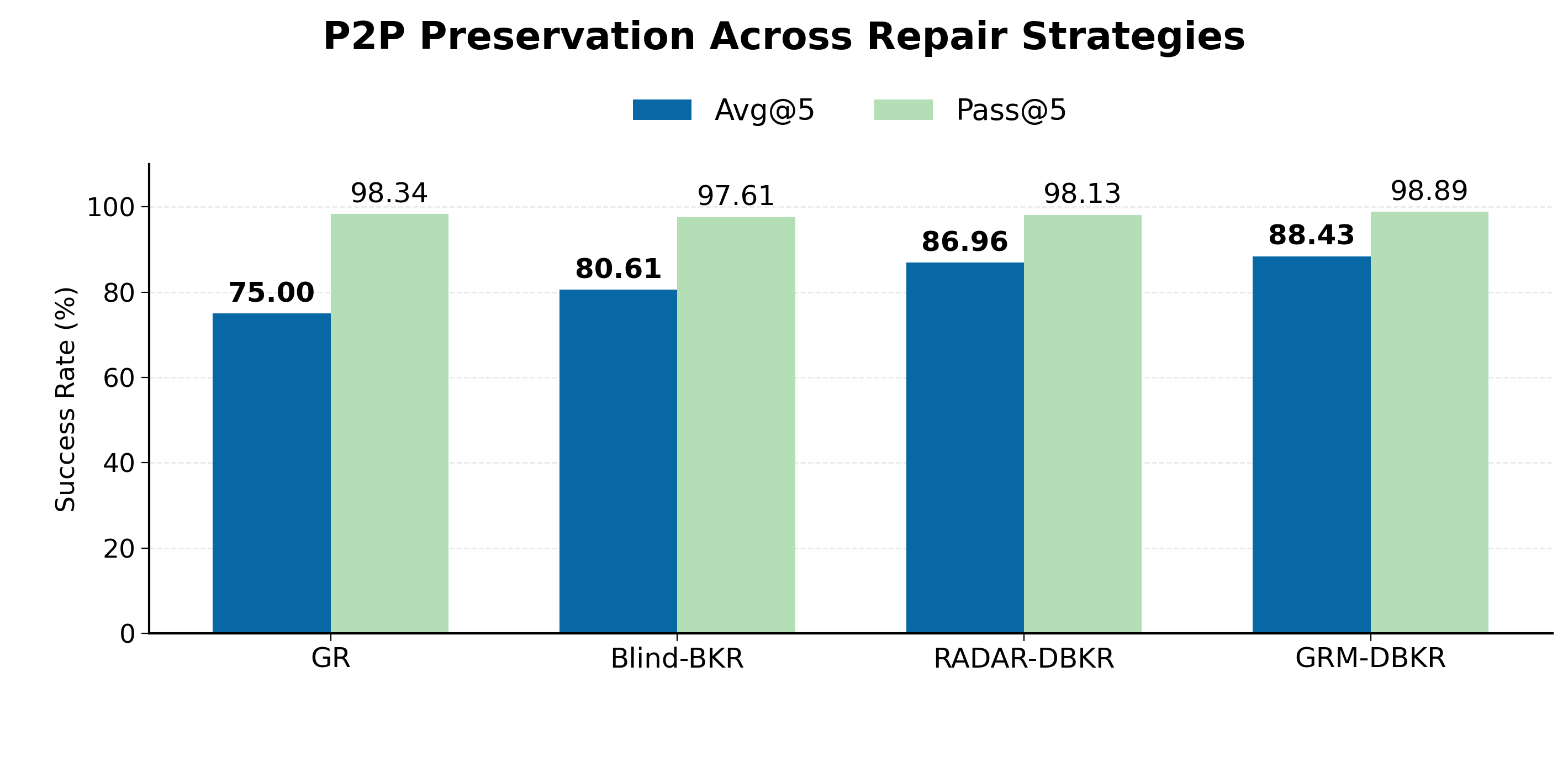}
        \caption{
            P2P preservation across RQ1 settings. Avg@5 measures mean continuation success; Pass@5 measures success in at least one of five continuations.
        }
        \label{fig:p2p_distribution}
    \end{minipage}
\end{figure}

\begin{figure*}[t]
\centering
\includegraphics[width=0.98\linewidth]{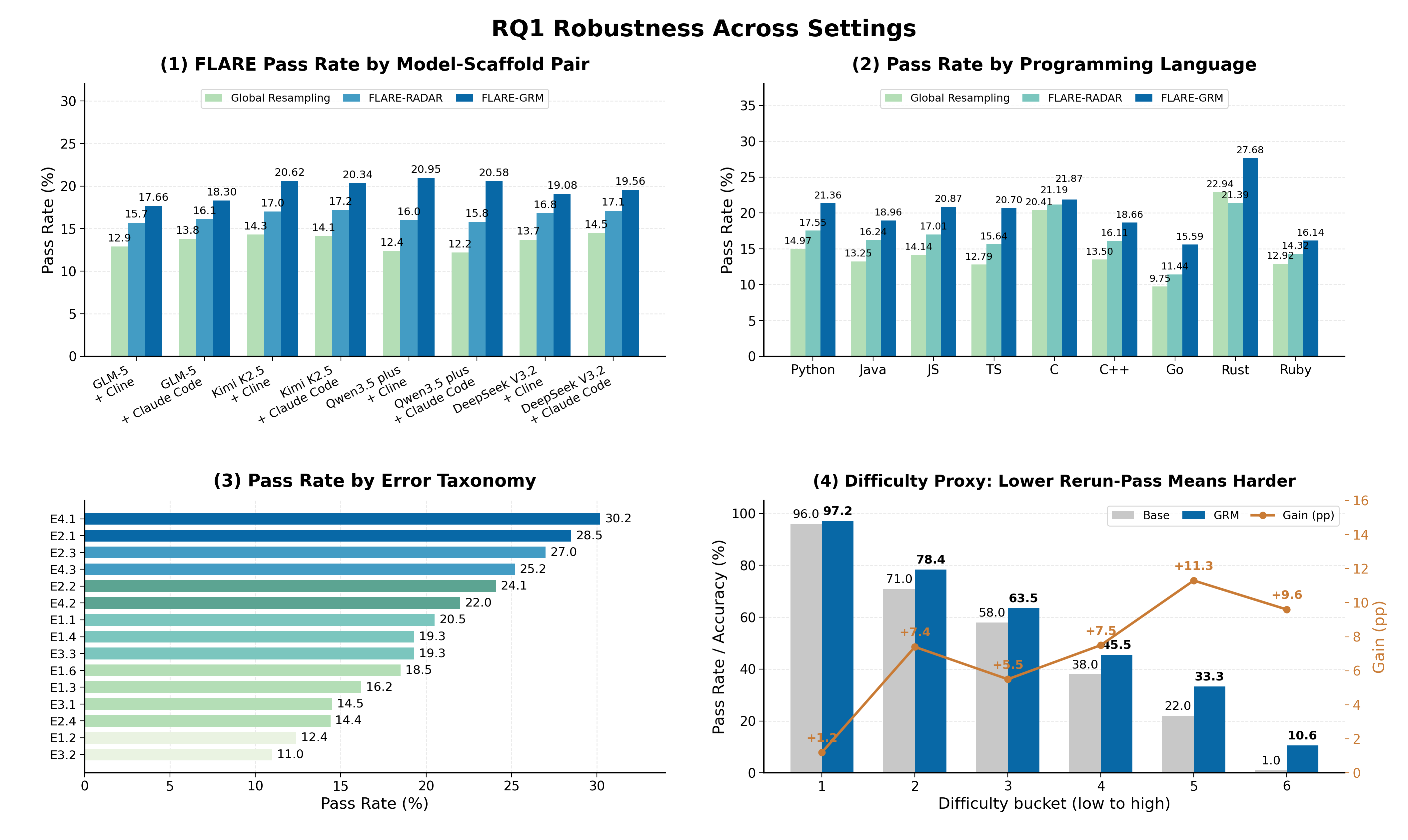}
\caption{
RQ1 robustness across settings. We report FLARE performance across model--scaffold combinations, programming languages, error taxonomy categories, and task-difficulty buckets. Across these slices, FLARE maintains positive gains over global resampling and RADAR-guided replay, with the largest improvements appearing in harder tasks where ordinary reruns rarely succeed.
}
\label{fig:rq1_robustness}
\end{figure*}

\textbf{Evaluation protocol.}
We split tasks by an initial baseline rollout, uniformly distributed across four models and two agent frameworks. Tasks that fail form the F2P subset, where success is measured by Pass@$N$ over repair branches. Tasks that pass form the P2P subset, where we report Avg@$N$ to measure whether interventions preserve successful behavior. We use $N=5$ as the maximum repair budget. This decomposition tests two complementary requirements for an intervention: rescuing failures and maintaining successful continuations. The resulting conditional rates diagnose intervention behavior rather than directly reporting full-benchmark solve rates. Table~\ref{tab:combined_results} and Figures~\ref{fig:f2p_curve}--\ref{fig:p2p_distribution} summarize the evaluated model--scaffold settings; Figure~\ref{fig:rq1_robustness} provides the corresponding slices.

\begin{table}[t]
\centering
\small
\setlength{\tabcolsep}{5.5pt}
\renewcommand{\arraystretch}{1.25}
\caption{
Combined test-time intervention performance across the RQ1 model--scaffold settings. $N$ denotes the repair-branch budget. Costs follow the accounting in Section~\ref{sec:exp_setup} and measure base-agent output, excluding agent inputs, GRM inputs/outputs, and latency. Subscripts denote relative changes from GR at the same $N$; \textcolor{deepgreen}{green} is favorable and \textcolor{deepred}{red} unfavorable.
}
\label{tab:combined_results}
\begin{tabular}{llccc}
\toprule
\textbf{Method} & \textbf{Budget} & \textbf{Score (\%)} & \textbf{Agent Output Tokens} & \textbf{Avg. Turns} \\
\midrule
\multicolumn{5}{c}{\cellcolor{gray!10}\textbf{Fail-to-Pass Subset} \quad [\textit{Score: F2P Pass@N}]} \\
\midrule
\multirow{2}{*}{GR}
& $N=1$ & 7.80 & \metricbarc{22738}{66711}{grgreen} & \metricbarc{103.9}{307.8}{grgreen} \\
& $N=5$ & 13.72 & \metricbarc{66711}{66711}{grgreen} & \metricbarc{307.8}{307.8}{grgreen} \\
\midrule
\multirow{2}{*}{Blind-BKR}
& $N=1$ & 0.50\baddn{93.59} & \metricbarc{19046}{66711}{blindgray} & \metricbarc{86.6}{307.8}{blindgray} \\
& $N=5$ & 1.20\baddn{91.25} & \metricbarc{48393}{66711}{blindgray} & \metricbarc{221.8}{307.8}{blindgray} \\
\midrule
\multirow{2}{*}{RADAR-DBKR (Ours)}
& $N=1$ & 10.89\pos{39.62} & \metricbarc{20814}{66711}{radarblue} & \metricbarc{92.8}{307.8}{radarblue} \\
& $N=5$ & 16.79\pos{22.38} & \metricbarc{55284}{66711}{radarblue} & \metricbarc{252.6}{307.8}{radarblue} \\
\midrule
\multirow{2}{*}{GRM-DBKR (Ours)}
& $N=1$ & \textbf{14.10}\pos{80.77} & \metricbarcB{12517}{66711}{grmblue} & \metricbarcB{57.6}{307.8}{grmblue} \\
& $N=5$ & \textbf{19.59}\pos{42.78} & \metricbarc{61718}{66711}{grmblue} & \metricbarc{284.2}{307.8}{grmblue} \\
\midrule
\multicolumn{5}{c}{\cellcolor{gray!10}\textbf{Pass-to-Pass Subset} \quad [\textit{Score: P2P Avg@N}]} \\
\midrule
GR
& $N=5$ & 75.00 & \metricbarc{10861}{10861}{grgreen} & \metricbarc{50.76}{50.76}{grgreen} \\
\cmidrule{1-5}
Blind-BKR
& $N=5$ & 80.61\pos{7.48} & \metricbarc{7836}{10861}{blindgray} & \metricbarc{37.60}{50.76}{blindgray} \\
RADAR-DBKR (Ours)
& $N=5$ & 86.96\pos{15.95} & \metricbarcB{7624}{10861}{radarblue} & \metricbarcB{35.94}{50.76}{radarblue} \\
GRM-DBKR (Ours)
& $N=5$ & \textbf{88.43}\pos{17.91} & \metricbarc{9400}{10861}{grmblue} & \metricbarc{48.15}{50.76}{grmblue} \\
\bottomrule
\end{tabular}
\end{table}

\textbf{F2P repair and efficiency.}
Table~\ref{tab:combined_results} shows that GRM-DBKR achieves the best repair rate at both budgets. With a single branch, it reaches 14.10\% F2P, nearly doubling GR at $N=1$ and exceeding the 13.72\% obtained by GR at $N=5$. The corresponding agent-output ratio is $66{,}711/12{,}517 \approx 5.33$, supporting the approximately $5\times$ reduction in base-agent output under this protocol. At $N=5$, GRM-DBKR further improves to 19.59\%, exceeding GR by 5.87 percentage points. Figure~\ref{fig:f2p_curve} shows that GRM-DBKR continues to benefit from additional branches over the evaluated range, while offline and blind replay methods saturate earlier. These results improve the observed repair-rate--agent-output trade-off.

\textbf{P2P preservation.}
On initially successful trajectories, GRM-DBKR obtains the highest P2P rate, 88.43\%, while reducing agent output relative to GR. Figure~\ref{fig:p2p_distribution} further shows that GRM-DBKR achieves the highest Avg@5 and Pass@5 among the evaluated strategies. This supports the stability of guided continuations under the evaluated protocol. A failed continuation is not itself evidence of GRM-induced harm, since P2P also reflects the stochasticity of the continuation policy.

\textbf{Why online intervention changes the cost-performance frontier.}
The headline numbers in Table~\ref{tab:combined_results} reflect a structural contrast in the timing of feedback. RADAR-DBKR waits for a completed trajectory to perform post-hoc attribution; GRM-DBKR can diagnose and redirect the first live trajectory. Global Rollout samples fresh trajectories without diagnostic feedback and does not intrinsically require a prior failure. The initial-rollout cost charged to GR in this experiment follows the conditional evaluation protocol, rather than an algorithmic requirement of global sampling.

At $N=1$, the reported protocol charges GR, Blind-BKR, and RADAR-DBKR for the baseline plus one repair branch, while GRM-DBKR uses its online-guided rollout. GRM-DBKR requires 12,517 agent-output tokens and 57.6 turns, compared with 22,738 tokens and 103.9 turns for GR, while increasing F2P from 7.80\% to 14.10\%. At $N=5$, GRM-DBKR uses 61,718 agent-output tokens versus GR's 66,711, with the 5.87-point F2P gain noted above. These counts reflect the actor's guided continuations, not tokens generated by the GRM. Online intervention changes when corrective computation is spent and reduces redundant actor generation in the evaluated setting.

\textbf{Robustness across settings.}
Figure~\ref{fig:rq1_robustness} further evaluates whether the RQ1 gains persist across model, scaffold, language, task-difficulty, and error-taxonomy slices. Gains are consistent across both Cline and Claude Code scaffolds and across GLM-5, Kimi K2.5, Qwen3.5-Plus, and DeepSeek V3.2, supporting the GRM's use as a plug-and-play intervention module across the evaluated settings. Across the nine evaluated language slices, FLARE yields positive absolute gains, including large lifts in TypeScript (+7.91 points) and Python (+6.39 points), with a positive gain also observed in C (+1.46 points).

The largest gains appear at the capability boundary. Using baseline rerun performance as a difficulty proxy, GRM-DBKR provides larger gains in the harder task buckets: in the lowest-baseline-success bucket, FLARE raises success from 1.0\% to 10.6\%. Taxonomy-level slicing also clarifies the method's operational scope. GRM-guided repair is strongest for process-level and validation failures, such as E4.1 Missing Validation (30.2\% repair) and E2.1 Missed Retrieval (28.5\% repair), but is less effective in the reported E3.2 and E1.2 slices. This pattern supports the intended interpretation: localized diagnosis is most useful when it redirects an otherwise viable repair process, while the final repair remains bounded by the base agent's capabilities.

\subsection{RQ2: Process-Aware Trajectory Scoring and SFT Data Curation}
\label{sec:rq2}

\textbf{Training-free score validation.}
Before using the FLARE process score for fine-tuning data selection, we test whether it aligns with ground-truth trajectory outcomes while keeping both the trained GRM and actor fixed. Here, \textit{training-free} refers to score evaluation, not to the construction of the GRM. Table~\ref{tab:auc_ablation_final} reports ROC-AUC over the 18,000-trajectory pool described in Section~\ref{sec:exp_setup}. The full FLARE Process-Aware score reaches 75.74\% ROC-AUC, outperforming chance ranking (50.00\%) and the heuristic Critical Model baseline (69.34\%). Aggregating step-level diagnostics therefore provides a useful ranking signal before downstream fine-tuning.

\begin{table}[t]
\centering
\small
\setlength{\tabcolsep}{4pt}
\renewcommand{\arraystretch}{1.12}
\caption{
Training-free leave-one-out ablation of the FLARE trajectory score. ROC-AUC is computed over the 18,000-trajectory pool with Pass/Fail outcomes, without updating the trained GRM or actor. Critical Model is a separate heuristic ranking baseline, not the rubric critic used in the SFT comparison. Subscripts denote absolute drops in percentage points from the full score.
}
\label{tab:auc_ablation_final}
\begin{tabularx}{\linewidth}{@{}Xccccc c@{}}
\toprule
\textbf{Scoring Rule}
&  $\omega_t^{risk}$
& $\omega_t^{err}$
& $\eta_t^{act}$
& $\eta_t^{rec}$
& $\eta_t^{pos}$
& \textbf{ROC-AUC (\%)} \\
\midrule
Random Sampling
& \textcolor{gray}{--} & \textcolor{gray}{--} & \textcolor{gray}{--} & \textcolor{gray}{--} & \textcolor{gray}{--}
& 50.00$_{\textcolor{gray}{\text{ }}}$ \\
Critical Model
& \textcolor{gray}{--} & \textcolor{gray}{--} & \textcolor{gray}{--} & \textcolor{gray}{--} & \textcolor{gray}{--}
& 69.34$_{\textcolor{gray}{\text{ }}}$ \\
FLARE Process-Aware (Ours)
& \textcolor{deepgreen}{\checkmark} & \textcolor{deepgreen}{\checkmark} & \textcolor{deepgreen}{\checkmark} & \textcolor{deepgreen}{\checkmark} & \textcolor{deepgreen}{\checkmark}
& 75.74$_{\textcolor{gray}{\text{ }}}$ \\
\midrule
w/o Risk Severity Factor ($\omega_t^{risk}$)
& \textcolor{deepred}{$\times$} & \textcolor{deepgreen}{\checkmark} & \textcolor{deepgreen}{\checkmark} & \textcolor{deepgreen}{\checkmark} & \textcolor{deepgreen}{\checkmark}
& 70.12$_{\textcolor{deepred}{\text{-5.62}}}$ \\
w/o Error Importance Factor ($\omega_t^{err}$)
& \textcolor{deepgreen}{\checkmark} & \textcolor{deepred}{$\times$} & \textcolor{deepgreen}{\checkmark} & \textcolor{deepgreen}{\checkmark} & \textcolor{deepgreen}{\checkmark}
& 71.06$_{\textcolor{deepred}{\text{-4.68}}}$ \\
w/o Action rarity weight ($\eta_t^{act}$)
& \textcolor{deepgreen}{\checkmark} & \textcolor{deepgreen}{\checkmark} & \textcolor{deepred}{$\times$} & \textcolor{deepgreen}{\checkmark} & \textcolor{deepgreen}{\checkmark}
& 72.95$_{\textcolor{deepred}{\text{-2.79}}}$ \\
w/o Recurrence penalty ($\eta_t^{rec}$)
& \textcolor{deepgreen}{\checkmark} & \textcolor{deepgreen}{\checkmark} & \textcolor{deepgreen}{\checkmark} & \textcolor{deepred}{$\times$} & \textcolor{deepgreen}{\checkmark}
& 73.20$_{\textcolor{deepred}{\text{-2.54}}}$ \\
w/o Position-aware weight ($\eta_t^{pos}$)
& \textcolor{deepgreen}{\checkmark} & \textcolor{deepgreen}{\checkmark} & \textcolor{deepgreen}{\checkmark} & \textcolor{deepgreen}{\checkmark} & \textcolor{deepred}{$\times$}
& 73.55$_{\textcolor{deepred}{\text{-2.19}}}$ \\
\bottomrule
\end{tabularx}
\end{table}

The leave-one-out ablation confirms that all five components contribute to predictive power. Consistent with the downstream SFT ablation below, the \textbf{Risk Severity Factor} ($\omega_t^{risk}$) and \textbf{Error Importance Factor} ($\omega_t^{err}$) are the most important intrinsic components: removing them causes the steepest absolute ROC-AUC drops, 5.62 and 4.68 points, respectively. The behavioral and temporal modifiers further refine the ranking among trajectories with similar diagnostic severity.

\begin{table*}[t]
\centering
\small
\setlength{\tabcolsep}{4pt}
\renewcommand{\arraystretch}{1.12}
\caption{
Combined SFT data-selection performance and scoring-component ablation. All methods fine-tune the Qwen3-30B-A3B base model with 3K selected trajectories. 
}
\label{tab:sft_combined}
\begin{tabularx}{\linewidth}{@{}Xccccc@{}}
\toprule
\textbf{Selection Strategy} & \textbf{Verified} & \textbf{Pro} & \textbf{Multilingual} & \textbf{Compass} & \textbf{Avg.} \\
\midrule
\multicolumn{6}{c}{\cellcolor{gray!10}\textbf{SFT Data Selection Strategies}} \\
\midrule
Random Sampling 
& 45.00$_{\textcolor{gray}{\text{ }}}$ 
& 20.11$_{\textcolor{gray}{\text{ }}}$ 
& 32.33$_{\textcolor{gray}{\text{ }}}$ 
& 18.00$_{\textcolor{gray}{\text{ }}}$ 
& 28.86$_{\textcolor{gray}{\text{ }}}$ \\

Rubric-Supervised Critic Selection 
& 47.00$_{\textcolor{deepgreen}{\text{+4.44}}}$ 
& 21.20$_{\textcolor{deepgreen}{\text{+5.42}}}$ 
& 33.67$_{\textcolor{deepgreen}{\text{+4.14}}}$ 
& 18.90$_{\textcolor{deepgreen}{\text{+5.00}}}$ 
& 30.19$_{\textcolor{deepgreen}{\text{+4.61}}}$ \\

FLARE Process-Aware (Ours) 
& 53.00$_{\textcolor{deepgreen}{\text{+17.78}}}$ 
& 24.90$_{\textcolor{deepgreen}{\text{+23.82}}}$ 
& 38.33$_{\textcolor{deepgreen}{\text{+18.56}}}$ 
& 21.30$_{\textcolor{deepgreen}{\text{+18.33}}}$ 
& 34.38$_{\textcolor{deepgreen}{\text{+19.13}}}$ \\

\midrule
\multicolumn{6}{c}{\cellcolor{gray!10}\textbf{Ablation of Process Score Components}} \\
\midrule
FLARE Process-Aware (Ours) 
& 53.00$_{\textcolor{gray}{\text{ }}}$ 
& 24.90$_{\textcolor{gray}{\text{ }}}$ 
& 38.33$_{\textcolor{gray}{\text{ }}}$ 
& 21.30$_{\textcolor{gray}{\text{ }}}$ 
& 34.38$_{\textcolor{gray}{\text{ }}}$ \\

\cmidrule{1-6}

w/o Risk Severity Factor ($\omega_t^{risk}$) 
& 49.20$_{\textcolor{deepred}{\text{-7.17}}}$ 
& 22.16$_{\textcolor{deepred}{\text{-11.00}}}$ 
& 35.00$_{\textcolor{deepred}{\text{-8.69}}}$ 
& 19.45$_{\textcolor{deepred}{\text{-8.69}}}$ 
& 31.45$_{\textcolor{deepred}{\text{-8.52}}}$ \\

w/o Error Importance Factor ($\omega_t^{err}$) 
& 49.60$_{\textcolor{deepred}{\text{-6.42}}}$ 
& 21.89$_{\textcolor{deepred}{\text{-12.09}}}$ 
& 36.33$_{\textcolor{deepred}{\text{-5.22}}}$ 
& 19.80$_{\textcolor{deepred}{\text{-7.04}}}$ 
& 31.91$_{\textcolor{deepred}{\text{-7.18}}}$ \\

w/o Action rarity weight ($\eta_t^{act}$) 
& 50.00$_{\textcolor{deepred}{\text{-5.66}}}$ 
& 23.26$_{\textcolor{deepred}{\text{-6.59}}}$ 
& 36.00$_{\textcolor{deepred}{\text{-6.08}}}$ 
& 19.95$_{\textcolor{deepred}{\text{-6.34}}}$ 
& 32.30$_{\textcolor{deepred}{\text{-6.05}}}$ \\

w/o Recurrence penalty ($\eta_t^{rec}$) 
& 51.60$_{\textcolor{deepred}{\text{-2.64}}}$ 
& 23.80$_{\textcolor{deepred}{\text{-4.42}}}$ 
& 37.00$_{\textcolor{deepred}{\text{-3.47}}}$ 
& 20.85$_{\textcolor{deepred}{\text{-2.11}}}$ 
& 33.31$_{\textcolor{deepred}{\text{-3.11}}}$ \\

w/o Position-aware weight ($\eta_t^{pos}$) 
& 50.80$_{\textcolor{deepred}{\text{-4.15}}}$ 
& 23.39$_{\textcolor{deepred}{\text{-6.06}}}$ 
& 37.33$_{\textcolor{deepred}{\text{-2.61}}}$ 
& 20.45$_{\textcolor{deepred}{\text{-3.99}}}$ 
& 32.99$_{\textcolor{deepred}{\text{-4.04}}}$ \\

\bottomrule
\end{tabularx}
\end{table*}

\textbf{SFT data selection.}
We select the top 3,000 trajectories from the distillation pool under each selection rule and fine-tune the same Qwen3-30B-A3B base model. As visualized in Figure~1, and shown in the top block of Table~\ref{tab:sft_combined}, FLARE Process-Aware Selection consistently outperforms random sampling and Rubric-Supervised Critic Selection across all four benchmarks. It improves the average pass rate by 19.13\% relative to random sampling, with especially large gains on SWE-bench Pro and SWE-Compass.

\textbf{Component ablation.}
The bottom block of Table~\ref{tab:sft_combined} shows that each score component contributes to downstream data quality. Removing the risk severity factor $\omega_t^{risk}$ or error importance factor $\omega_t^{err}$ causes the largest average SFT drops, 8.52\% and 7.18\%, respectively. The action-rarity, recurrence, and position-aware terms provide smaller but consistent gains by refining the ranking among trajectories with similar diagnostic severity. This mirrors the training-free ROC-AUC evidence in Table~\ref{tab:auc_ablation_final}: the same two diagnostic factors are also the most important for ranking passing trajectories above failing ones before any fine-tuning occurs.

\vspace{0.5em}
\noindent\textbf{Takeaway for RQ2:}
Structured process-aware scoring selects stronger SFT demonstrations than the evaluated random and rubric-critic baselines, with consistent gains across four benchmarks and supporting component ablations.

\vspace{-0.1cm}
\subsection{RQ3: Dense Reward for Agentic Reinforcement Learning}
\label{sec:rq3}

\textbf{Protocol.}
RQ3 tests whether the same GRM signal remains useful when used as an RL reward. We compare sparse-reward RL, which uses only the final repository-level Pass/Fail outcome, with FLARE-RL, which augments policy optimization with GRM-derived dense rewards. Both variants use the same model initialization, data source, rollout budget, optimizer, learning-rate schedule, and evaluation harness; only the reward construction differs. Step rewards are aggregated before computing the shared trajectory advantage, following Section~\ref{sec:grm}.

\begin{table}[t]
\centering
\small
\setlength{\tabcolsep}{4pt}
\caption{
RQ3: RL comparison between outcome-only sparse rewards and GRM-derived dense rewards under the shared initialization and training protocol in Section~\ref{sec:exp_setup}. Scores are pass rates on the four evaluation benchmarks; Avg.\ is their unweighted mean.
}
\label{tab:rl_main}
\begin{tabular}{lccccc}
\toprule
\textbf{Reward Signal} 
& \textbf{Verified} 
& \textbf{Pro} 
& \textbf{Multilingual} 
& \textbf{Compass} 
& \textbf{Avg.} \\
\midrule
Sparse outcome reward
& 57.40$_{\textcolor{gray}{\text{ }}}$
& 27.63$_{\textcolor{gray}{\text{ }}}$
& 42.00$_{\textcolor{gray}{\text{ }}}$
& 24.05$_{\textcolor{gray}{\text{ }}}$
& 37.77$_{\textcolor{gray}{\text{ }}}$ \\

Dense FLARE reward (Ours)
& \textbf{60.20}$_{\textcolor{deepgreen}{\text{+4.88}}}$
& \textbf{31.60}$_{\textcolor{deepgreen}{\text{+14.37}}}$
& \textbf{44.67}$_{\textcolor{deepgreen}{\text{+6.36}}}$
& \textbf{28.50}$_{\textcolor{deepgreen}{\text{+18.50}}}$
& \textbf{41.24}$_{\textcolor{deepgreen}{\text{+9.19}}}$ \\
\bottomrule
\end{tabular}
\end{table}

\textbf{Results.}
Table~\ref{tab:rl_main} shows that FLARE-RL improves over sparse-reward RL on all four benchmarks: +2.80 points on SWE-bench Verified, +3.97 on SWE-bench Pro, +2.67 on SWE-bench Multilingual, and +4.45 on SWE-Compass. Since the two RL variants differ only in the reward signal, this consistent improvement supports dense process supervision as a useful complement to terminal correctness, even when optimization uses a shared trajectory-level advantage.

\vspace{0.5em}
\noindent\textbf{Takeaway for RQ3:}
The GRM signal transfers from offline trajectory ranking to online policy optimization, completing the lifecycle use of FLARE across inference, SFT, and RL.

\vspace{-0.2cm}
\section{Conclusion}
FLARE unifies test-time inference and train-time alignment through structured GRM diagnostics, connecting causal-aware supervision construction, online intervention, and dense reward learning in long-horizon SWE tasks. Under the conditional repair protocol, one GRM-guided branch achieves higher F2P than five global reruns while reducing base-agent output by approximately $5\times$, improving the observed repair-rate--agent-output frontier. Reusing the same diagnostic signal for SFT selection and dense RL rewards delivers relative performance gains of 19.13\% and 9.19\%, respectively. Together, these results establish a unified framework for turning trajectory diagnosis into actionable supervision across the agent lifecycle.
\bibliographystyle{plainnat}
\bibliography{references}

\clearpage
\appendix

\lstdefinestyle{appendixcode}{
    basicstyle=\ttfamily\footnotesize\linespread{1.08}\selectfont,
    keywordstyle=\ttfamily,
    commentstyle=\ttfamily\itshape,
    stringstyle=\ttfamily,
    columns=fullflexible,
    keepspaces=true,
    showstringspaces=false,
    breaklines=true,
    breakatwhitespace=true,
    frame=single,
    framerule=0.3pt,
    rulecolor=\color{black!25},
    backgroundcolor=\color{black!2},
    xleftmargin=1em,
    xrightmargin=1em,
    aboveskip=0.75em,
    belowskip=0.75em,
    tabsize=4
}

\lstdefinestyle{appendixprompt}{
    basicstyle=\ttfamily\footnotesize\linespread{1.05}\selectfont,
    keywordstyle=\ttfamily,
    columns=fullflexible,
    keepspaces=true,
    showstringspaces=false,
    breaklines=true,
    breakatwhitespace=true,
    frame=single,
    framerule=0.3pt,
    rulecolor=\color{black!25},
    backgroundcolor=\color{black!2},
    xleftmargin=1em,
    xrightmargin=1em,
    aboveskip=0.8em,
    belowskip=0.8em,
    tabsize=2
}

\definecolor{promptbg}{RGB}{248,249,252}
\definecolor{promptframe}{RGB}{48,113,190}

\newtcblisting{promptbox}[1]{%
  listing only,
  colback=promptbg,
  colframe=promptframe,
  title={\small\bfseries #1},
  fonttitle=\small\bfseries,
  colbacktitle=promptframe,
  coltitle=white,
  listing options={%
    basicstyle=\ttfamily\scriptsize,
    breaklines=true,
    breakautoindent=false,
    columns=fullflexible,
    keepspaces=true,
    showstringspaces=false,
    aboveskip=0pt,
    belowskip=0pt,
  },
  boxrule=0.6pt,
  arc=2pt,
  left=6pt,
  right=6pt,
  top=2pt,
  bottom=2pt,
  breakable,
}

\tcbset{
  promptstyle/.style={
    listing only,
    colback=promptbg,
    colframe=promptframe,
    colbacktitle=promptframe,
    coltitle=white,
    fonttitle=\small\bfseries,
    listing options={
      basicstyle=\ttfamily\scriptsize,
      breaklines=true,
      breakautoindent=false,
      columns=fullflexible,
      keepspaces=true,
      showstringspaces=false,
      aboveskip=0pt,
      belowskip=0pt,
    },
    boxrule=0.6pt,
    arc=2pt,
    left=6pt,
    right=6pt,
    top=2pt,
    bottom=2pt,
    breakable,
  },
}
\section{Limitations}
While FLARE establishes a highly effective dense supervision paradigm for long-horizon SWE tasks, we acknowledge several limitations that provide directions for future research:

Bounded by Base Agent Semantic Capabilities. FLARE functions as an alignment and intervention scaffold rather than a substitute for fundamental reasoning. As observed during test-time intervention, while the Generative Reward Model (GRM) excels at localizing process-level missteps and validation failures, its effectiveness diminishes when failures require deep semantic architectural shifts or when the agent experiences severe goal drift. If the base model lacks the inherent coding capability to implement a complex repair, simply pinpointing the error step is insufficient to rescue the trajectory.

Overhead of Step-Level Generative Diagnosis. Although GRM-DBKR substantially reduces base-agent output under the reported repair protocol, diagnostic inference incurs additional computation and potential latency. The reported token counts measure agent outputs only; end-to-end resource use also depends on agent inputs, GRM inference, environment execution, and the serving configuration. The one-time costs of offline annotation and GRM training are also outside this accounting.

Dependency on a Predefined Error Taxonomy. RADAR uses a fixed taxonomy to make diagnostic supervision interpretable and reusable, but this design may miss unfamiliar errors outside the annotated categories. We audited diagnostic severity, evidence grounding, and leakage through independent review by human SWE experts and a stronger language model, with disagreements adjudicated against the visible prefix (Appendix~\ref{app:radar_prompts}). This audit measures residual annotation errors in the sampled diagnostics rather than guaranteeing error-free supervision.

\section{Error Taxonomy}
\label{app:taxonomy}

This appendix provides the fixed error taxonomy used throughout the RADAR--GRM pipeline, including offline RADAR annotation, causal-prefix distillation, runtime GRM diagnosis, GRM-guided checkpoint repair, and taxonomy-level experimental analysis. The taxonomy is designed to characterize actionable process-level failures in SWE-agent trajectories, rather than merely describe the final Pass/Fail outcome. All non-safe RADAR annotations and GRM risk diagnoses are constrained to use only rubric codes from this taxonomy.

\newcommand{\catheader}[1]{%
\midrule
\multicolumn{3}{c}{\textbf{#1}} \\
\midrule
}

\begin{longtable}{L{0.12\linewidth} L{0.28\linewidth} L{0.52\linewidth}}
\caption{Error taxonomy used for RADAR annotation and GRM step-level diagnosis.}
\label{tab:error_taxonomy}\\
\toprule
\textbf{Code} & \textbf{Error Type} & \textbf{Definition / Evidence Cue} \\
\midrule
\endfirsthead

\toprule
\textbf{Code} & \textbf{Error Type} & \textbf{Definition / Evidence Cue} \\
\midrule
\endhead

\bottomrule
\endfoot

\catheader{C1 Task Understanding \& Planning}

E1.1 & Misunderstood Requirements &
The agent misinterprets the task goal or ignores explicit constraints; modifies the wrong target or bypasses the required interface. \\

E1.2 & Goal Drift &
The agent drifts from the original goal to irrelevant objectives such as unrelated refactoring, formatting, or premature completion. \\

E1.3 & Incorrect Step Decomposition &
The agent omits necessary inspection, setup, dependency, or validation steps, causing failures from skipped prerequisites. \\

E1.4 & Wrong Tool Selection &
The agent uses an inappropriate or fragile tool when a direct, reliable, or project-native method is available. \\

E1.5 & Ordering / Dependency Error &
The agent executes steps in an invalid order, such as using files before generation, running before setup, or committing before validation. \\

E1.6 & Incomplete Implementation / Residual TODOs &
The agent claims completion while leaving placeholders, TODOs, missing logic, or partial call-chain updates. \\

\noalign{\penalty-10000}
\catheader{C2 Context \& Evidence Handling}

E2.1 & Irrelevant or Missed Retrieval &
The agent searches the wrong scope or misses relevant files, symbols, implementation locations, or evidence needed for the task. \\

E2.2 & Wrong Evidence Selected &
Relevant evidence is available, but the agent selects the wrong file, snippet, candidate, or code path. \\

E2.3 & Misinterpreted Tool Output &
The agent misreads compiler, test, linter, shell, or diagnostic output and takes an ineffective action. \\

E2.4 & Lost Context / Memory Mismatch &
The agent forgets or mismatches earlier requirements, constraints, or established conclusions in a long trajectory. \\

\catheader{C3 Code Generation \& Modification}

E3.1 & Syntax / Type Error &
The generated or modified code violates syntax, compilation, interpretation, or static type-checking constraints. \\

E3.2 & API / Library Misuse or Version Mismatch &
The agent misuses an API, library, framework convention, or version-specific behavior. \\

E3.3 & Semantic / Logic Error &
The code runs or appears plausible but violates the required behavior due to incorrect algorithmic, boundary, business-logic, or state reasoning. \\

\catheader{C4 Validation \& Self-Check}

E4.1 & Missing Validation / Insufficient Coverage &
The agent skips necessary tests or validates too narrowly after non-trivial changes. \\

E4.2 & Shallow Check &
The agent only checks that execution succeeds, without verifying outputs, side effects, or edge cases. \\

E4.3 & Ignored Failure Signal / Misread Test Report &
The agent ignores or dismisses clear failure signals from tools, tests, logs, or CI. \\

\end{longtable}

\section{RADAR Annotation and Quality-Control Prompts}
\label{app:radar_prompts}

This appendix provides the prompts used by RADAR to construct causally grounded step-level supervision before GRM training. These offline prompts are distinct from the runtime GRM prompt in Appendix~\ref{app:grm_prompt}. The pipeline contains differential causal backtracking, causal-prefix distillation, and semantic consistency checking. Full-trajectory evidence supports offline attribution, while online explanations are constrained to the task and the prefix through the current step. The checks enforce this information boundary at the prompt level rather than establishing a formal guarantee of zero leakage.

We complemented the automated gate with an independent audit of 608 diagnostics by 3 human SWE experts and a stronger language model, GPT-5 (\texttt{gpt-5-2025-08-07}). Diagnostics were sampled according to stratified random sampling by repository and predicted severity, with one expert assigned to each diagnostic. Human and model reviewers independently assessed each diagnostic using only the task and current prefix, marking severity errors, unsupported repair advice, patch leakage, and future- or test-oracle information. Disagreements were adjudicated against the visible evidence. Table~\ref{tab:diagnostic_audit} reports adjudicated error counts and rates separately from downstream repair performance. Pre-adjudication agreement on severity labels was 92.60\%, with unweighted Cohen's $\kappa=0.887$ between the assigned expert and the model. The audit provides an empirical check of the information boundary in addition to automated prompt compliance.

\begin{table}[htbp]
\centering
\small
\caption{Independent diagnostic audit. Entries report error count / eligible diagnostics (error rate in percent) after adjudication. False-critical errors use diagnostics labeled critical as the denominator; the other categories use all audited diagnostics. Categories may overlap.}
\label{tab:diagnostic_audit}
\begin{tabularx}{\linewidth}{>{\raggedright\arraybackslash}X l}
\toprule
\textbf{Audit criterion} & \textbf{Adjudicated errors} \\
\midrule
Incorrect severity assignment & 37 / 608 (6.09\%) \\
False critical prediction & 14 / 163 (8.59\%) \\
Unsupported repair advice & 23 / 608 (3.78\%) \\
Patch leakage & 3 / 608 (0.49\%) \\
Future- or test-oracle information & 7 / 608 (1.15\%) \\
\bottomrule
\end{tabularx}
\end{table}

\subsection{Offline Differential Causal Backtracking and Joint Step-level Attribution Prompt}
\label{app:radar_offline_prompt}

\begin{promptbox}{Error attribution prompt}
# RADAR Offline Differential Causal Backtracking and Joint Step-level Attribution Prompt

You are the offline diagnostic module of RADAR. Your task is to perform post-hoc causal analysis over SWE-agent trajectories and construct step-level supervision data for GRM training.

Read all inputs from the predefined directory:

- problem_statement.txt: original task requirement.
- traj_summary.txt: final result claimed by the agent.
- failing_tests.txt: final failure signals.
- golden_patch.patch: reference patch, used only for offline diagnosis.
- model_patch.patch: agent-produced patch.
- abstract_llm.json: compressed per-turn trajectory with intent, action, and observation.
- trajectory.json: raw trajectory, used only when abstract_llm.json is insufficient.
- rubrics.jsonl: fixed error taxonomy. Do not invent or modify rubric categories.

================================================================
Stage A: Differential Causal Backtracking
================================================================

Construct a causal error chain explaining how final failure, patch discrepancy, or requirement violation arose from earlier trajectory decisions.

The chain must be:

- plausible: every node helps explain the final failure or deviation;
- complete: major causally relevant decisions are captured;
- precise: unrelated, merely redundant, stylistic, or inefficient turns are excluded.

Procedure:

1. Difference anchoring.
   Compare golden_patch.patch and model_patch.patch semantically. Identify affected files, symbols, interfaces, behaviors, contracts, and how each mismatch relates to failing_tests.txt or the task requirement. Patch and failing-test information is allowed only in offline_analysis and must not appear in later online_analysis.

2. Backward traversal.
   Traverse abstract_llm.json backward. Mark a turn as a causal-chain node only if concrete evidence shows that it:
   - directly created, modified, validated, or preserved a behavior in the final deviation; or
   - introduced or committed to an assumption, contract interpretation, scope decision, or evidence selection required for the final deviation.
   Discard incidental, exploratory, redundant, or unrelated turns.

3. Root-cause priority.
   Identify the earliest observable turn where the erroneous assumption, wrong contract, wrong evidence selection, or invalid implementation direction first appears. This is the preferred root-cause breakpoint. Later propagation or amplification turns may be non-safe, but they are not preferred over the root cause for breakpoint replay.

4. Independent roots.
   If multiple failures arise from independent assumptions or modifications, output separate root nodes.

Each causal-chain node must contain:

{
  "turn_id": <int>,
  "role": "root" | "propagation" | "amplification",
  "divergence_anchor": "<file/symbol/interface/contract/behavior>",
  "dependency": <turn_id | null>
}

where root is the earliest source, propagation carries forward a prior error, amplification expands or solidifies a prior error, and dependency points to the earlier causal node or null for an independent root.

================================================================
Stage B: Joint Step-level Attribution
================================================================

Annotate every turn in abstract_llm.json with exactly one record containing:

1. risk_level
2. rubric_hits
3. offline_analysis

risk_level must be one of:

- safe: the turn is correct, neutral, or constructively aligned with the task.
- trivial: the turn has a real defect, but it is unlikely to independently cause final failure or major derailment.
- critical: the turn can independently cause task failure, substantially derail the trajectory, validate a wrong assumption, or make recovery significantly harder.

Severity must be determined by the current turn's causal role and evidence, not by a fixed rubric whitelist. Any rubric category may be critical if the causal role is critical. Do not force any target label distribution.

For every non-safe turn, assign one or more rubric hits from rubrics.jsonl:

{
  "rule_id": "<rubric code>",
  "label": "<canonical label>",
  "confidence": "high" | "low",
  "reason": "<specific evidence from the current turn and relevant context>"
}

Rules:

- safe turns must have rubric_hits = [].
- non-safe turns must have at least one rubric hit.
- every rubric code and label must exist in rubrics.jsonl.
- do not assign rubrics merely because the final outcome failed.

offline_analysis must be one concise English paragraph per turn.

For safe turns, explain what the turn did and why it is aligned or neutral.

For non-safe turns, explain:

1. what the turn did;
2. what went wrong, with evidence;
3. how it relates to final failure, patch discrepancy, or requirement violation;
4. why the severity is appropriate;
5. the minimal corrective direction.

offline_analysis may use golden_patch.patch, model_patch.patch, failing_tests.txt, and future turns because it is post-hoc. Such hindsight information must later be removed by causal-prefix distillation before GRM training or online repair.

================================================================
Consistency Requirements
================================================================

- Every turn in abstract_llm.json must have exactly one annotation.
- Every causal-chain node must have risk_level in {"trivial", "critical"}.
- A turn outside the causal chain may be non-safe only for a local defect unrelated to the final failure; offline_analysis must state this.
- If a later critical turn is merely downstream of an earlier safe/trivial turn, revisit and update the earlier turn before finalizing.
- Do not label a turn non-safe solely for verbosity, redundancy, or inefficiency unless it creates concrete task risk.
- Do not force any safe/trivial/critical ratio.

================================================================
Output
================================================================

Return exactly one parseable JSON object and nothing else:

{
  "causal_chain": [
    {
      "turn_id": <int>,
      "role": "root" | "propagation" | "amplification",
      "divergence_anchor": "<string>",
      "dependency": <int | null>
    }
  ],
  "annotations": [
    {
      "turn_id": <int>,
      "risk_level": "safe" | "trivial" | "critical",
      "rubric_hits": [
        {
          "rule_id": "<string>",
          "label": "<string>",
          "confidence": "high" | "low",
          "reason": "<string>"
        }
      ],
      "offline_analysis": "<English paragraph>"
    }
  ]
}

Do not output Markdown, comments, explanations, or code fences.
\end{promptbox}

\subsection{Causal-prefix Distillation Prompt}
\label{app:causal_prefix_prompt}

\begin{promptbox}{Error attribution prompt}
# RADAR Causal-prefix Distillation Prompt

You are the causal-prefix distillation module of RADAR. Your task is to rewrite post-hoc offline_analysis into online_analysis using only information visible up to and including the current turn. The resulting online_analysis is used for GRM training and may be used as checkpoint-repair feedback.

This prompt constructs training annotations. It is distinct from the runtime GRM prompt, which may use compact parseable outputs such as <safe/>.

Read all inputs from the predefined directory:

- annotated_trajectory.json: output of offline RADAR attribution, including turn_id, risk_level, rubric_hits, and offline_analysis.
- problem_statement.txt: original task requirement, always visible to the downstream agent.
- abstract_llm.json: compressed trajectory. For each turn, only the prefix up to and including that turn is visible.
- trajectory.json: raw trajectory, used only to verify prefix-visible anchors.
- rubrics.jsonl: fixed taxonomy, used only to preserve rubric meaning. Do not invent categories.

================================================================
Hard Causal Masking Constraints
================================================================

Every online_analysis must satisfy all constraints below:

1. No patch leakage.
   Do not mention or reveal golden_patch.patch, model_patch.patch, diff hunks, reference/expected/correct patch content, or patch-only paths, symbols, constants, or implementation details not visible in the prefix.

2. No final-evaluation leakage.
   Do not mention final failing_tests.txt, final evaluation results, final patch outcome, future turns, or facts visible only after the current turn. Mentions of test failures are allowed only if the failure message is directly visible in the current or earlier trajectory prefix.

3. No test-oracle leakage.
   Do not mention specific test function names, hidden expected values, assertions, fixtures, or symbols that appear only inside test files. Test directories or test file paths may be mentioned only if already prefix-visible and used solely as validation targets.

4. Prefix-only evidence.
   Use only problem_statement.txt, the current turn, and earlier turns from abstract_llm.json or trajectory.json.

5. English only.

6. No unsupported anchors.
   Do not invent symbols, paths, line numbers, APIs, constants, commands, outputs, or error messages. Every concrete anchor must be prefix-visible or stated in problem_statement.txt.

7. No unsupported runtime/library claims.
   Do not assert runtime, standard-library, framework, or dependency semantics as decisive unless directly visible in the prefix. Otherwise, phrase the issue using observed code, command output, or task requirements.

8. No speculative wording for confirmed non-safe issues.
   For trivial or critical turns, avoid hedges such as may, might, possibly, could potentially, perhaps, seems to, appears to. If evidence is insufficient, weaken the severity instead.

================================================================
Content Requirements
================================================================

Append exactly one field to each annotation:

"online_analysis": "<English paragraph or compact structured English text>"

Preserve the input causal_chain unchanged in the output.

For safe turns:

- write 2--3 concise English sentences;
- describe what the turn did;
- explain why it is aligned, neutral, or useful;
- end with a natural sentence stating the risk level.

For trivial or critical turns, include all five components:

1. Current action:
   State what the turn did and cite at least one current-turn prefix-visible anchor, such as a path, symbol, branch condition, command, error message, or tool output.

2. What remains on track:
   Acknowledge useful or correct partial progress so replay does not discard valid context.

3. Decisive issue:
   State 1--3 concrete issues. Each must cite a prefix-visible anchor, explain conflict with the problem statement or visible code/output, avoid patch/future/test-oracle information, and use direct non-speculative wording.

4. Remaining:
   Include an explicit field:

   Remaining:
   - <requirement>

   or:

   Remaining: none

   Items must come from problem_statement.txt and prefix-visible implementation status, not hidden tests or patches.

5. Suggested next step:
   Give the minimal concrete next action: which existing file/function/symbol/module to inspect or modify, what direction to take, and how to keep the change within scope. If a relevant test path is prefix-visible, mention only that path as a validation target. End with a natural sentence stating the risk level.

================================================================
Output
================================================================

Return exactly one parseable JSON object and nothing else:

{
  "causal_chain": [...],
  "annotations": [
    {
      "turn_id": <int>,
      "risk_level": "safe" | "trivial" | "critical",
      "rubric_hits": [...],
      "offline_analysis": "<original offline analysis>",
      "online_analysis": "<causally masked English analysis>"
    }
  ]
}

Do not output Markdown, comments, explanations, or code fences.
\end{promptbox}

\subsection{Semantic Consistency and Causal Compliance Gate Prompt}
\label{app:semantic_gate_prompt}

\begin{promptbox}{Error attribution prompt}
# RADAR Semantic Consistency and Causal Compliance Gate Prompt

You are the quality-control gate for RADAR annotations. Audit a fully annotated trajectory before it is used for GRM training, checkpoint repair, trajectory scoring, or reward construction. Check logical consistency, causal consistency, taxonomy consistency, prefix-visibility compliance, and output completeness. Do not create new annotations unless explicitly requested.

Read all inputs from the predefined directory:

- annotated_trajectory.json: contains causal_chain and annotations with turn_id, risk_level, rubric_hits, offline_analysis, and online_analysis.
- problem_statement.txt: original task requirement.
- abstract_llm.json: compressed trajectory.
- trajectory.json: raw trajectory, used only to verify prefix-visible evidence.
- rubrics.jsonl: fixed taxonomy.

================================================================
D1: Logical and Taxonomy Consistency
================================================================

Check that:

- safe turns have rubric_hits = [].
- trivial/critical turns have non-empty rubric_hits.
- every rubric_hit.rule_id exists in rubrics.jsonl.
- every rubric_hit.label matches rubrics.jsonl.
- every rubric_hit.reason cites concrete evidence from the corresponding turn or necessary prefix context.
- risk_level is justified by causal role, not by a rubric whitelist.
- any rubric category may be critical if the step can cause failure, derail the trajectory, validate a wrong assumption, or make recovery harder.
- trivial does not describe a clearly decisive failure unless offline_analysis explains why it is recoverable or non-decisive.
- critical does not rely only on final failure; it must identify a current-turn defect or decision.

================================================================
D2: Causal-chain Consistency
================================================================

Check that:

- every causal_chain node maps to exactly one annotation.
- every causal_chain node has risk_level in {"trivial", "critical"}.
- every divergence_anchor is concrete and evidence-supported.
- root nodes have dependency = null.
- propagation/amplification nodes have valid earlier dependencies unless justified as independent roots.
- no later critical turn is merely the downstream consequence of an earlier safe turn.
- if a later critical turn depends on an earlier trivial turn, offline_analysis must justify why the earlier turn is recoverable or less severe.
- non-safe turns outside the causal chain must be explicitly described as local non-causal defects in offline_analysis.

================================================================
D3: Online-analysis Causal Masking Compliance
================================================================

For every online_analysis, check:

- no patch leakage: no golden/reference/expected/correct patch, diff hunk, or prefix-invisible patch-only details.
- no future-outcome leakage: no future turns, final evaluation, final failing_tests.txt, final patch outcome, or facts visible only after the current turn.
- test-failure mentions are allowed only if directly visible in the current or earlier prefix.
- no test-oracle leakage: no specific test function names, hidden expected values, assertions, fixtures, or test-only symbols.
- test directories/files are allowed only if prefix-visible and used only as validation targets.
- English only.
- every non-safe online_analysis cites at least one prefix-visible or problem-statement anchor.
- no invented paths, line numbers, symbols, APIs, constants, commands, outputs, or error messages.
- trivial/critical issues avoid speculative wording such as may, might, possibly, could potentially, perhaps, seems to, appears to.
- every trivial/critical online_analysis contains "Remaining:" with items or "Remaining: none".
- every safe online_analysis ends with a natural sentence stating the risk level.

================================================================
D4: Coverage and Format Completeness
================================================================

Check that:

- every turn in abstract_llm.json has exactly one annotation.
- annotations contain no duplicate turn_id.
- every annotation includes turn_id, risk_level, rubric_hits, offline_analysis, and online_analysis.
- causal_chain and annotations fields both exist.
- the artifact is valid parseable JSON.

================================================================
Violation Severity and Routing
================================================================

Use "block" if the artifact must be regenerated before downstream use, including patch leakage, future leakage, invalid JSON, missing/duplicate annotation, invalid rubric code, safe with rubric_hits, non-safe without rubric_hits, causal-chain node labeled safe, unsupported critical label, or non-English online_analysis.

Use "warn" for non-blocking issues such as mild vagueness, redundancy, overly verbose safe analysis without leakage, or weak but parseable validation reminders.

Routing:

- causal_chain, risk_level, rubric_hits, or offline_analysis issues route to "RADAR_attribution".
- online_analysis or causal masking issues route to "causal_prefix_distillation".
- include both routes if both issue types occur.

================================================================
Output
================================================================

Return exactly one parseable JSON object and nothing else:

{
  "passed": true | false,
  "violations": [
    {
      "turn_id": <int | null>,
      "dimension": "D1" | "D2" | "D3" | "D4",
      "rule": "<short rule identifier>",
      "evidence": "<quoted or summarized evidence>",
      "severity": "block" | "warn",
      "route_back_to": "RADAR_attribution" | "causal_prefix_distillation"
    }
  ],
  "route_back_to": [
    "RADAR_attribution",
    "causal_prefix_distillation"
  ]
}

If there are no violations, return:

{
  "passed": true,
  "violations": [],
  "route_back_to": []
}

Do not output Markdown, comments, explanations, or code fences.
\end{promptbox}

\section{Runtime GRM Prompt}
\label{app:grm_prompt}

This appendix gives the exact runtime prompting format used by the GRM to evaluate a single step during SWE-agent execution. Unlike the RADAR prompts in Appendix~\ref{app:radar_prompts}, which construct and validate training annotations under offline or distilled supervision settings, this prompt is used at inference time. The GRM receives the problem statement, prefix trajectory, current step, rubric taxonomy, and tool descriptions, and returns either a safe judgment or a rubric-backed risk judgment in a parseable XML format. This output format enables the same diagnostic signal to be reused for checkpoint repair, trajectory scoring, and dense reward construction.

\subsection{System Prompt}
\label{app:grm_system_prompt}

\begin{promptbox}{Error attribution prompt}
You are a Generative Reward Model (GRM) that evaluates exactly one current step in a SWE agent trajectory.

## Input Structure

The user message will always use these exact XML tags:

- <problem_statement>: the task description the agent is solving.
- <history_trajectory>: all prior steps before the current one, in raw JSON format. Each step is an assistant message (which may contain reasoning and tool calls) followed by the corresponding tool result messages.
- <current_step_to_evaluate>: the single current step to judge, in the same raw JSON format.
- <rubrics>: the error taxonomy. Each entry contains a code, definition, criteria, and examples.
- <tools>: descriptions of all tools available to the agent in its framework.

## Your Task

Judge whether <current_step_to_evaluate> contains one or more rubric-backed errors.

- Use <problem_statement>, <history_trajectory>, and <tools> only as context.
- Evaluate only the current step itself.
- Only mention a prior-step mistake if the current step explicitly repeats it or builds on it.
- Use only rubric codes that appear in <rubrics>.

## Hard Output Contract

Return exactly one top-level XML result and nothing else.

### Option A: safe

<safe/>

### Option B: risk

<risk>
    <analysis>
        {what went wrong, why it matters, and what concrete evidence in the current step supports the judgment}
    </analysis>
    <issue rubric="{code}" label="{level_1_category}/{level_2_label}"/>
    <severity>{critical|trivial}</severity>
</risk>

If there are multiple rubric-backed errors, output multiple <issue .../> lines inside the same <risk> block.

## Required Rules

- The first character of your response must be `<`.
- The last character of your response must be `>` from the final XML tag.
- Do not output any natural language before the XML block.
- Do not output any natural language after the XML block.
- Do not output Markdown, code fences, JSON, comments, XML declarations, or any `<think>` / `</think>` tags.
- Never output both `<safe/>` and `<risk>` in the same response.
- If the step is safe, output exactly `<safe/>` and stop.
- If the step is risky, output exactly one top-level `<risk>` block.
- A `<risk>` response must contain:
  - exactly one `<analysis> ... </analysis>` block,
  - at least one `<issue .../>` line,
  - exactly one `<severity>` whose value is either `critical` or `trivial`.
- Do not emit any other XML tags.
- Escape reserved XML characters in text and attribute values (for example, & as &amp; and < as &lt;).

## Decision Rules

- `critical`: the issue is likely to cause task failure, a wrong final result, or a major derailment.
- `trivial`: the issue exists, but is unlikely to affect final correctness or overall progress.
- If any listed issue is critical, the final severity must be `critical`.
- If you cannot support an error with concrete evidence from the current step plus context, prefer `<safe/>`.

## Correct / Incorrect Output Examples

Correct:
<safe/>

Incorrect:
<risk>
    <analysis>
        ...
    </analysis>
    <issue rubric="E1.2" label="Task Understanding &amp; Planning Errors/Goal Drift"/>
    <severity>trivial</severity>
</risk>
\end{promptbox}

\subsection{User Prompt Template}
\label{app:grm_user_prompt}

\begin{promptbox}{Error attribution prompt}
<problem_statement>
{problem_statement}
</problem_statement>

<history_trajectory>
{history_trajectory}
</history_trajectory>

<current_step_to_evaluate>
{current_step_to_evaluate}
</current_step_to_evaluate>

<rubrics>
{rubrics}
</rubrics>

<tools>
{tools}
</tools>
\end{promptbox}

\section{Case Study: GRM-Guided Checkpoint Rollout}
\label{app:case_study}

We provide a qualitative case study to illustrate how FLARE uses GRM feedback to recover a long-horizon SWE trajectory through localized checkpoint repair. This case is intended as an interpretability example rather than additional quantitative evidence. Unlike aggregate F2P metrics, the goal here is to expose the concrete mechanism by which a GRM intervention changes the subsequent implementation path.

\paragraph{Task.}
The task is from \texttt{alibaba/fescar\#644}. The original issue reports a serialization failure in the distributed transaction rollback path:
\path{java.io.NotSerializableException: com.alibaba.fescar.rm.datasource.sql.struct.Null}.
The failing object is \path{com.alibaba.fescar.rm.datasource.sql.struct.Null}, a sentinel object used to represent SQL \texttt{NULL} values in prepared-statement parameters. Therefore, the task is not merely to add a marker interface somewhere in the codebase. A correct repair must make \texttt{Null} serializable while preserving its intended sentinel-object semantics.

\paragraph{Why this case is informative.}
This trajectory contains a particularly subtle failure mode. Before the GRM interruption, the agent had already found the right file and had made a superficially plausible edit: it modified \texttt{Null.java} so that \texttt{Null} implements \texttt{java.io.Serializable}. However, this only addresses the surface exception. The class is also exposed through \texttt{Null.get()}, indicating that it is used as a singleton-style sentinel object. Under that design, serialization must not create semantically distinct \texttt{Null} objects after deserialization. The GRM intervention is useful because it prevents the agent from treating ``implements \texttt{Serializable}'' as a complete solution.

\subsection{Breakpoint Context}
\label{app:case_study_breakpoint_context}

At the interruption point, the agent had narrowed the problem to the correct class and stated the following implementation intent:

\begin{lstlisting}[style=appendixprompt]
The `Null` class is used to represent SQL NULL values in prepared statements. When Fescar performs distributed transaction rollbacks, it needs to serialize these objects, but `Null` doesn't implement `Serializable`.

The fix is to make the `Null` class implement `java.io.Serializable`:
\end{lstlisting}

The immediately following tool output shows that the actual edit only added the serialization marker interface and a \texttt{serialVersionUID} field:

\begin{lstlisting}[style=appendixcode]
package com.alibaba.fescar.rm.datasource.sql.struct;

import java.io.Serializable;

/**
 * The type Null.
 */
public class Null implements Serializable {

    private static final long serialVersionUID = 748997787189430077L;
\end{lstlisting}

This is the critical breakpoint. The agent is on the right file and has addressed the exception type, but the current patch has not yet handled the semantic interaction between Java serialization and the singleton sentinel pattern.

\subsection{GRM Diagnosis}
\label{app:case_study_grm_output}

Given the current step and its prefix context, the GRM produced the following critical-risk diagnosis:

\begin{lstlisting}[style=appendixprompt]
<risk>
    <analysis>
        The agent edits Null.java to add import java.io.Serializable, implements Serializable, and serialVersionUID field. The current action modifies the Null.java file. What is on track: The agent correctly identifies that Null class needs to be serializable for distributed transaction rollback scenarios. Decisive issue: The agent implements Serializable but does not consider the class's singleton pattern (instance field) and toString() method. The existing code shows Null.get() returns a singleton instance; making it serializable may not align with the singleton design. The agent also does not examine how Null is used in prepared statements (AbstractPreparedStatementProxy.java) to ensure serialization works correctly. The implementation is incomplete for the task requirements. Remaining: 1) Ensure Null class is serializable for distributed transaction rollback scenarios, 2) Verify Null.get() singleton pattern works with serialization, 3) Check prepared statement usage of Null in AbstractPreparedStatementProxy.java for serialization compatibility. Suggested next step: Examine AbstractPreparedStatementProxy.java to see how Null.get() is used in setParamByIndex, then adjust Null.java to maintain singleton pattern while making it serializable. Run tests in rm-datasource module before declaring done. The turn is critical.
    </analysis>
    <issue rubric="E3.3" label="Code Generation &amp; Modification Errors/Semantic / Logic Error (algorithm / boundary condition / state)"/>
    <issue rubric="E1.6" label="Task Understanding &amp; Planning Errors/Incomplete Implementation / Residual TODOs"/>
    <severity>critical</severity>
</risk>
\end{lstlisting}

This diagnosis is important for two reasons. First, it does not discard the useful part of the agent's work: the GRM explicitly recognizes that making \texttt{Null} serializable is directionally correct for the rollback failure. Second, it identifies the missing semantic condition: the fix must preserve the \texttt{Null.get()} singleton pattern across serialization boundaries. The assigned taxonomy labels also match the failure mode: the current edit is a semantic logic error (\texttt{E3.3}) and an incomplete implementation (\texttt{E1.6}), rather than a syntax or tool-use failure.

\subsection{Checkpoint Replay and Behavioral Change}
\label{app:case_study_replay}

After checkpoint replay, the agent's next steps directly followed the GRM's repair direction. Instead of immediately declaring the task complete, the agent first re-examined the current implementation and explicitly reframed the problem around the singleton pattern:

\begin{lstlisting}[style=appendixprompt]
Let me examine the current state of the file and understand the singleton pattern issue:
\end{lstlisting}

The replayed branch then articulated the missing failure mechanism:

\begin{lstlisting}[style=appendixprompt]
Now I understand the usage. The `Null.get()` singleton is used to represent SQL NULL values in prepared statement parameters. The issue is that when serializable objects containing the `Null` instance are serialized and then deserialized, Java's default serialization will create a new object, breaking the singleton pattern.

To fix this properly, I need to add a `readResolve()` method to ensure the singleton pattern is preserved during deserialization:
\end{lstlisting}

The resulting code-level repair adds a \texttt{readResolve()} method to \texttt{Null.java}:

\begin{lstlisting}[style=appendixcode]
private Object readResolve() {
    return instance;
}
\end{lstlisting}

This small change is the key evidence that the replayed branch internalized the GRM feedback. The original interrupted branch treated serializability as a marker-interface requirement. The repaired branch instead implements singleton-safe serialization: \texttt{Null} can now participate in serialized rollback state, while deserialization resolves back to the canonical singleton instance used by \texttt{Null.get()}.

\subsection{Effect of the Intervention}
\label{app:case_study_effect}

Table~\ref{tab:case_study_fescar} summarizes the causal role of this single GRM interruption.

\begin{table}[H]
\centering
\small
\caption{GRM-guided checkpoint repair in the \texttt{alibaba/fescar\#644} case. The GRM prevents a shallow serializability patch from being treated as complete and redirects the replay toward singleton-safe serialization.}
\label{tab:case_study_fescar}
\begin{tabularx}{\linewidth}{L{0.22\linewidth} L{0.34\linewidth} >{\raggedright\arraybackslash}X}
\toprule
\textbf{Stage} & \textbf{Observed behavior} & \textbf{Interpretation} \\
\midrule
Before interruption &
The agent edits \texttt{Null.java} to add \texttt{import java.io.Serializable}, \texttt{implements Serializable}, and \texttt{serialVersionUID}. &
The agent has found the right class and addressed the surface exception, but the implementation is incomplete because it does not preserve the singleton sentinel semantics of \texttt{Null.get()}. \\

GRM diagnosis &
The GRM marks the step as \texttt{critical}, citing \texttt{E3.3} and \texttt{E1.6}. It states that the implementation must verify the singleton pattern and inspect prepared-statement usage through \texttt{AbstractPreparedStatementProxy.java}. &
The feedback is localized and actionable: it keeps the correct target file, preserves the useful partial edit, and identifies the missing semantic contract rather than requesting a full restart. \\

After checkpoint replay &
The agent re-examines the singleton pattern, reasons that default deserialization would create a new object, and adds \texttt{readResolve()} returning \texttt{instance}. &
The repaired branch changes the completion criterion from ``\texttt{Null} implements \texttt{Serializable}'' to ``\texttt{Null} remains the canonical SQL NULL sentinel across serialization and deserialization.'' \\
\bottomrule
\end{tabularx}
\end{table}
\FloatBarrier

\paragraph{Discussion.}
This case illustrates the advantage of GRM-guided repair over Blind-BKR. The diagnosed step is not obviously catastrophic: it has already modified the correct file and implemented the interface named by the exception; the intervention redirects the subsequent continuation. However, it is still a high-risk process step because it creates a plausible but shallow patch that can prematurely terminate the trajectory. The GRM feedback exposes the hidden semantic requirement, namely preserving the singleton-style \texttt{Null} sentinel across serialization. Checkpoint replay then uses this diagnosis to produce a more precise repair in \texttt{Null.java}. In this way, FLARE does not merely resample a new continuation; it converts a step-level diagnosis into a concrete implementation constraint that changes the subsequent behavior of the agent.

\end{document}